\documentclass[a4paper,fleqn]{cas-sc}
\usepackage[utf8]{inputenc}
\usepackage[authoryear]{natbib}
\usepackage{lineno}
\usepackage{siunitx}    
\usepackage{algorithmic}
\usepackage{orcidlink}
\usepackage{subcaption}
\usepackage{tabularx}
\usepackage{caption}
\usepackage{tikz}
\usepackage{ragged2e}
\usepackage{placeins}

\definecolor{cb1st}{RGB}{86,180,233}   
\definecolor{cb2nd}{RGB}{230,159,0}    
\definecolor{cb3rd}{RGB}{0,158,115}    

\graphicspath{ {./figures/} }
\usetikzlibrary{shapes.geometric, arrows.meta, positioning, fit, backgrounds, calc, decorations.pathreplacing}

\begin{document}
\nolinenumbers

\shorttitle{Evaluating ML Models for Post-Wildfire Debris-Flow Prediction}
\shortauthors{Q. Ledingham et al.}

\title[mode=title]{Evaluating Machine Learning Models for Post-Wildfire Debris-Flow Prediction}

\author[1]{Quinn Ledingham}
\author[1]{Zhengsen Xu}
\author[1]{Yimin Zhu}
\author[1]{Zack Dewis}
\author[1]{Mabel Heffring}
\author[1]{Saeid Taleghanidoozdoozan}
\author[1]{Motasem Alkayid}
\author[1]{Megan Greenwood}
\author[1]{Lincoln Linlin Xu}
\cormark[1]
\ead{lincoln.xu@ucalgary.ca}

\affiliation[1]{organization={Department of Geomatics Engineering, University of Calgary},
                city={Calgary},
                state={AB},
                country={Canada}}

\cortext[1]{Corresponding author}

\begin{abstract}
Prediction of post-wildfire debris flows is critical for mitigating hazards to communities, infrastructure, and resources during intense rainfall in recently burned areas.
However, identifying reliable machine learning models for this task is complicated by several factors, including the overlap of debris-flow and no-debris-flow events in feature space, the lack of physical interpretability, and limited training data.
Therefore, the goal of this paper is to address these challenges through a systematic evaluation of a broad set of machine learning models in terms of model performance, feature importance, and response to synthetic data augmentation.
Using basin-scale observations of post-wildfire debris-flow events across the western United States, we compare 15 models, including a new foundation model, the Tabular Prior-Data Fitted Network (TabPFN).
The results of repeated stratified cross-validation indicate that TabPFN achieves the strongest unaugmented performance with a threat score of 0.637, closely followed by the leading tree-based models.
We use SHapley Additive exPlanations (SHAP) for the feature importance evaluation to identify which features the top-performing models rely on most when predicting debris flows.
This evaluation finds that short-duration rainfall intensity and storm accumulation receive the highest SHAP rankings across the explained models, with burn severity and terrain features ranked lower.
Finally, we quantify the utility of synthetic data augmentation to mitigate the scarcity of debris-flow event observations.
Using TabPFN to generate synthetic training data, we improve the performance of all models except CNN, with the largest mean gain in threat score of $+0.041$ among the deep-learning models.
By integrating rigorous model benchmarking, transparent feature evaluation, and data augmentation, this work provides a comprehensive framework for enhancing the accuracy and reliability of post-wildfire debris-flow prediction.
\end{abstract}


\maketitle

\section{Introduction}

\begin{figure}
    \includegraphics[width=\linewidth]{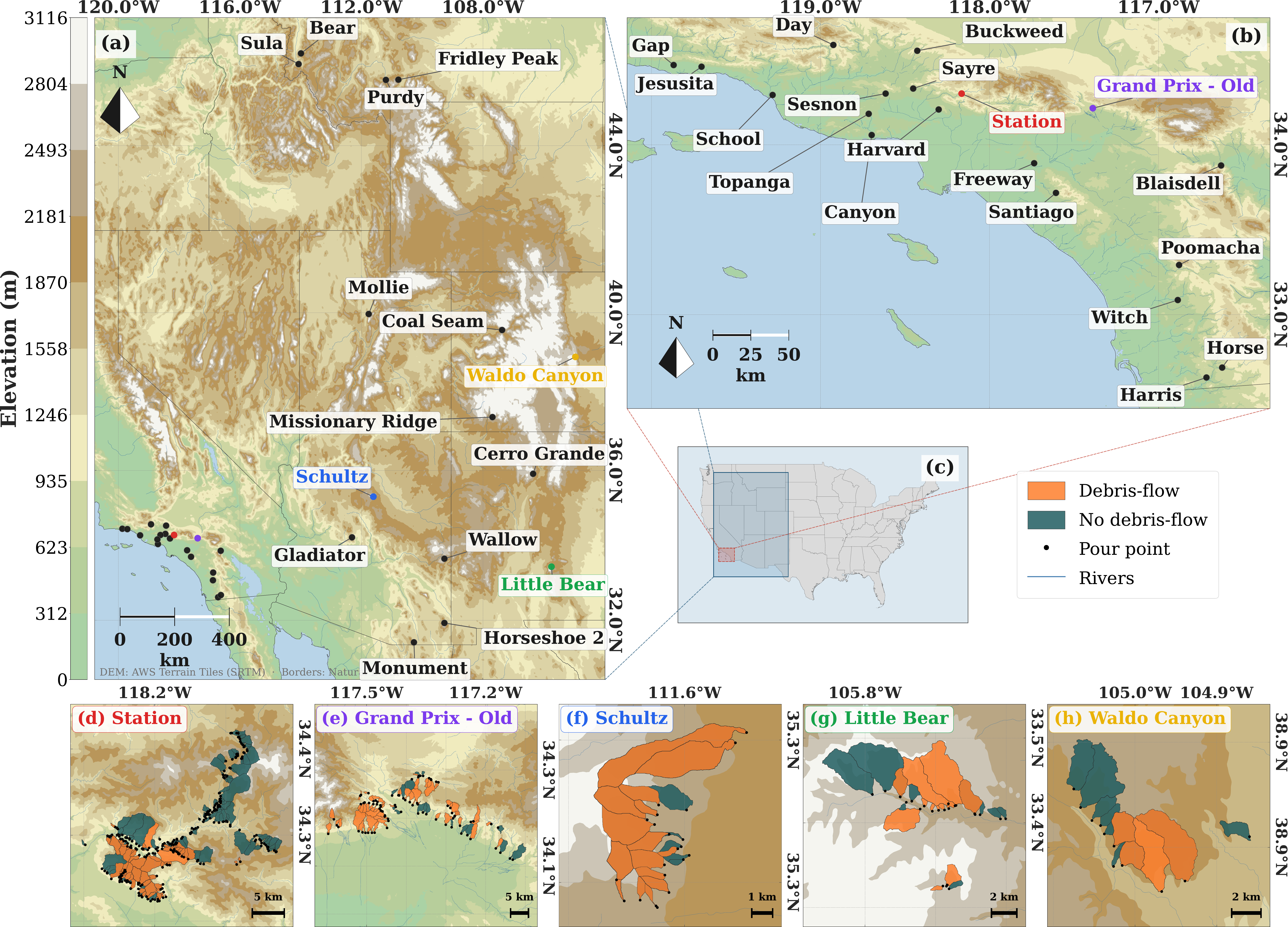}
    \caption{Study-area overview of the 34 fire events in the USGS post-wildfire debris-flow dataset (2000--2013, western United States).
    Panel~(a) shows the regional extent with labeled fire centroids on a digital elevation model (DEM), panel~(b) zooms to the southern California cluster, and panel~(c) shows a United States locator.
    Panels~(d)--(h) show delineated basins for the Station, Grand Prix--Old, Schultz, Little Bear, and Waldo Canyon fires, with polygons filled to show the response (no-debris-flow or debris flow).}
    \label{fig:events_map}
\end{figure}

Post-wildfire debris flows threaten lives, infrastructure, and resources in burned watersheds across fire-prone regions worldwide, and operational warning depends on models that estimate the probability of initiation from basin and storm features \citep{staley_prediction_2017}.
The problem of predicting this hazard is difficult for three reasons.
First, selecting the best model for this task is challenging.
Many candidate models are available, and debris-flow and non-event basins overlap in their rainfall, burn-severity, and terrain signatures (Table~\ref{tab:dataset_summary}; Figure~\ref{fig:tsne_tabpfn_oof}), so no single threshold cleanly separates the two classes of outcomes \citep{nikolopoulos_evaluation_2018}, making the choice of model consequential.
Second, when classes overlap in feature space, different model families can place decision boundaries differently, so identifying which features a high-performing model actually relies on is non-trivial.
This feature importance is needed to support model-selection and trust judgments, to diagnose whether predictions reflect robust signal rather than spurious patterns confined to a particular subset of observations, and to assess whether a model would generalize \citep{ribeiro_why_2016}.
Third, the available data for this problem are typically small and class-imbalanced, with debris-flow events the minority outcome, which constrains how well any model can learn the boundary between the two outcomes and disproportionately limits identification of the rare positive class \citep{he_learning_2009}.
Together, these difficulties motivate a rigorous, multi-model evaluation under controlled conditions, similar to the strategies used in other specialized remote-sensing prediction studies \citep{xu_comparative_2014}.

A typical post-wildfire debris-flow hazard assessment workflow proceeds from delineation of burned basins to assembly of rainfall and terrain features as a tabular collection of features to prediction of debris flows \citep{staley_updated_2016}.
Within this workflow, the prediction stage is critical because missed events leave exposed communities without warning, whereas false alarms increase operational burden \citep{sattele_reliability_2015}.
This paper, therefore, focuses on the prediction stage, where input features are translated into warning decisions.
Accordingly, we evaluate models using event-detection metrics, such as threat score, that balance missed events and false alarms under class imbalance, rather than overall accuracy alone \citep{schaefer_critical_1990}.

Several machine learning models have been applied to post-wildfire debris-flow prediction.
Logistic regression models that relate rainfall intensity and duration to initiation probability underlie current operational approaches by the United States Geological Survey (USGS) \citep{staley_objective_2013, staley_updated_2016, staley_prediction_2017}.
Based on the USGS approach, improved threat scores have been reported for naïve Bayes, mixture discriminant analysis, and classification trees \citep{kern_machine_2017}, decision trees \citep{addison_assessment_2019}, eXtreme Gradient Boosting (XGBoost) trained on satellite-based inputs \citep{orland_scalable_2022}, and Random Forest and neural network models \citep{nikolopoulos_evaluation_2018, roten_machine_2022}.
Beyond the United States, logistic regression models have been applied to post-wildfire debris-flow prediction in the Mediterranean and southwest China \citep{diakakis_exploring_2023, jin_susceptibility_2022}.
Three gaps remain in this body of work.
First, prior studies typically compare only a handful of models and rely on a single train-test split.
This makes performance estimates highly variable and heavily dependent on the specific split, particularly when performance gaps are small \citep{cawley_over-fitting_2010, krstajic_cross-validation_2014, dietterich_approximate_1998}.
In addition, several high-performing recently developed models for tabular data remain unevaluated for this problem.
Tabular Prior-Data Fitted Network (TabPFN) is particularly relevant, as it performs approximate Bayesian prediction via in-context learning to excel on small-to-medium datasets \citep{hollmann_tabpfn_2023, hollmann_accurate_2025}.
Beyond the debris-flow setting, TabPFN has outperformed conventional models in thaw-hazard mapping \citep{zhu_method_2026} and has shown strong performance for landslide susceptibility mapping under data-limited conditions \citep{zhou_landslide_2025, yang_knowledge-data_2026}.
Similarly, the tree-structured Recursive Feature Machine (xRFM) provides a scalable approach to feature learning that has shown strong performance on tabular benchmarks \citep{beaglehole_xrfm_2025}.
Second, the features that high-performing models for post-wildfire debris-flow prediction rely on are not well characterized.
A meta-analysis of debris-flow machine learning studies confirmed that interpretation methods are rarely applied across this literature \citep{yang_machine-learning-based_2024}.
Relatedly, whether different model families attend to the same features or to divergent signals remains an open question.
Third, while synthetic data augmentation has improved related hazard models under similar class imbalance \citep{wang_optimizing_2019, kumar_addressing_2024}, whether such augmentation transfers to the post-wildfire debris-flow setting, and which models benefit when it is added, is not well understood.

Therefore, this paper presents a rigorous evaluation framework for post-wildfire debris-flow prediction.
We conduct a systematic comparison of fifteen models across five model families: logistic, distance- and kernel-based, deep-learning, tree-based, and tabular foundation models.
Each model is trained under a shared preprocessing pipeline and a common hyperparameter-tuning framework where applicable, then evaluated with repeated stratified $k$-fold cross-validation.
This isolates model architecture as the primary variable and averages performance over many resamplings, so observed gaps reflect model choice rather than uneven configuration effort or sampling noise from a single split.
Beyond model ranking, we use SHapley Additive exPlanations (SHAP) feature importances \citep{lundberg_unified_2017} both to characterize the features that individual high-performing models rely on and to contrast importance patterns across model families on shared splits, asking whether high-ranked models converge on the same features or rely on divergent signals.
Finally, we generate synthetic training observations with TabPFN and quantify, under the same evaluation protocol, whether augmentation improves prediction in this class-imbalanced setting and which of the fifteen models possibly benefit from it.
Together, these results clarify which model families best handle the overlap between debris-flow and non-event basins, identify which features high-performing models rely on and where families agree or diverge, and quantify whether synthetic augmentation can mitigate the small, class-imbalanced training data typical of this task.

\section{Methodology}
\label{methodology}


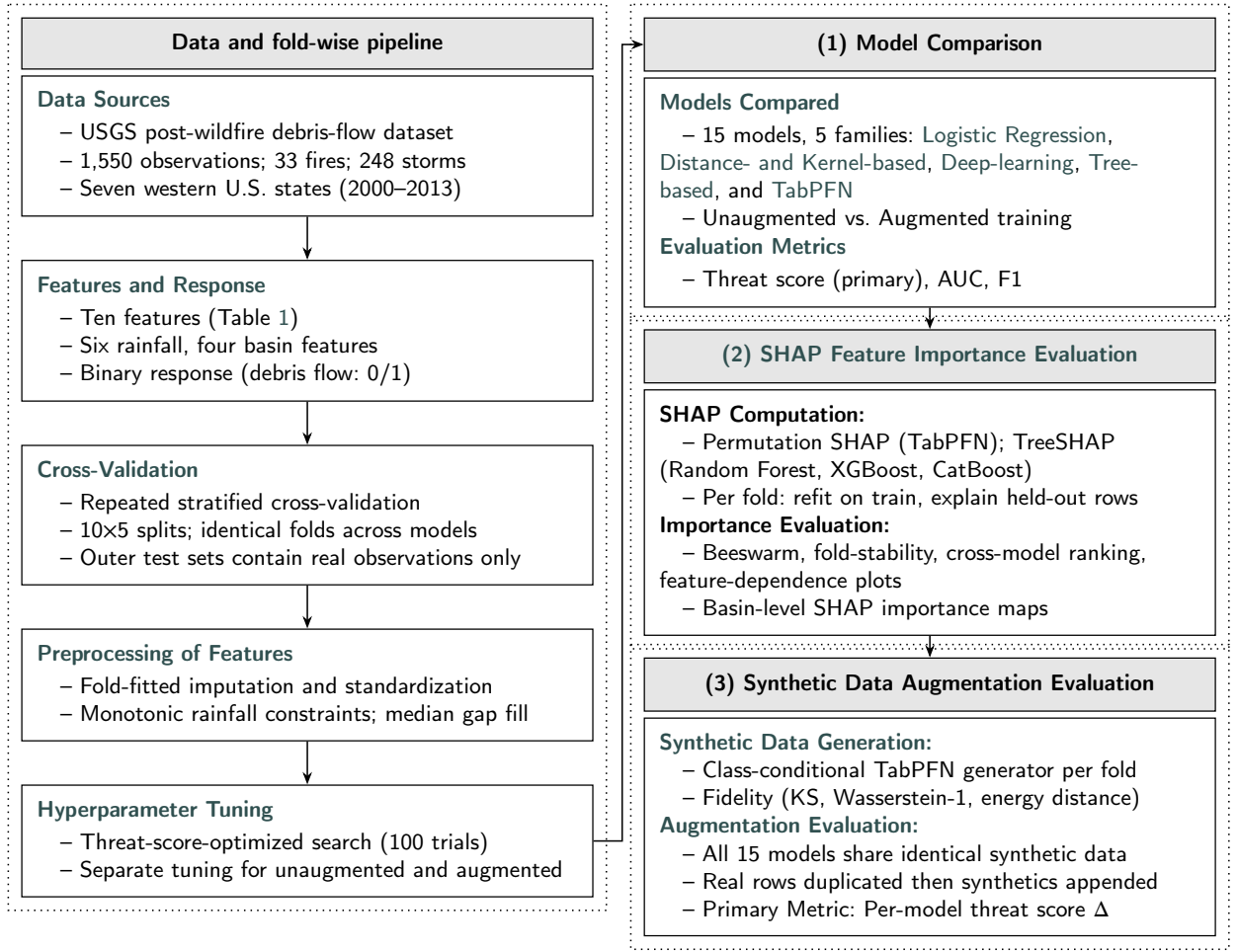
\begin{figure}
\centering
\resizebox{\textwidth}{!}{%
\begin{tikzpicture}[
    W/.style={minimum width=7.8cm, text width=7.53cm},
    colhdr/.style={
        W,
        rectangle, draw=black, line width=0.5pt,
        fill=gray!20,
        minimum height=0.6cm,
        align=center,
        inner sep=6pt,
        font=\bfseries\small
    },
    secbox/.style={
        W,
        rectangle, draw=black, line width=0.5pt,
        fill=white,
        align=left,
        inner sep=6pt,
        font=\small
    },
    arr/.style={-{Stealth[length=5pt]}, black, line width=0.6pt}
]

\node[colhdr, anchor=north west] (Lhdr) at (0,0) {Data and fold-wise pipeline};

\node[secbox, anchor=north west] (L1) at ($(Lhdr.south west)+(0,-0.08)$) {%
    \hyperref[dataset-data-sources]{\textbf{Data Sources}}\\[2pt]
    \quad -- USGS post-wildfire debris-flow dataset\\
    \quad -- 1{,}550 observations; 33 fires; 248 storms\\
    \quad -- Seven western U.S.\ states (2000--2013)%
};

\node[secbox, anchor=north west] (L2) at ($(L1.south west)+(0,-0.6)$) {%
    \hyperref[dataset-features-response]{\textbf{Features and Response}}\\[2pt]
    \quad -- Ten features (Table~\ref{tab:used_features})\\
    \quad -- Six rainfall, four basin features\\
    \quad -- Binary response (debris flow: 0/1)%
};

\node[secbox, anchor=north west] (L3) at ($(L2.south west)+(0,-0.6)$) {%
    \hyperref[cross-validation-section]{\textbf{Cross-Validation}}\\[2pt]
    \quad -- Repeated stratified cross-validation\\
    \quad -- 10$\times$5 splits; identical folds across models\\
    \quad -- Outer test sets contain real observations only%
};

\node[secbox, anchor=north west] (L4) at ($(L3.south west)+(0,-0.6)$) {%
    \hyperref[preprocessing-section]{\textbf{Preprocessing of Features}}\\[2pt]
    \quad -- Fold-fitted imputation and standardization\\
    \quad -- Monotonic rainfall constraints; median gap fill%
};

\node[secbox, anchor=north west] (L5) at ($(L4.south west)+(0,-0.6)$) {%
    \hyperref[hyperparameter-tuning-section]{\textbf{Hyperparameter Tuning}}\\[2pt]
    \quad -- Threat-score-optimized search (100 trials)\\
    \quad -- Separate tuning for unaugmented and augmented%
};

\draw[arr] (L1) -- (L2);
\draw[arr] (L2) -- (L3);
\draw[arr] (L3) -- (L4);
\draw[arr] (L4) -- (L5);

\coordinate (Rorigin) at ($(Lhdr.north east)+(0.7cm,0)$);

\node[colhdr, anchor=north west] (Rhdr) at (Rorigin) {(1) Model Comparison};

\node[secbox, anchor=north west] (R1) at ($(Rhdr.south west)+(0,-0.08)$) {%
    \hyperref[models-compared-section]{\textbf{Models Compared}}\\[2pt]
    \quad -- 15 models, 5 families: \hyperref[logistic-regression-section]{Logistic Regression}, \hyperref[distance-kernel-section]{Distance- and Kernel-based},  \hyperref[deep-learning-section]{Deep-learning}, \hyperref[tree-based-section]{Tree-based}, and \hyperref[tabpfn-section]{TabPFN}\\
    \quad -- Unaugmented vs.\ Augmented training\\
    \hyperref[evaluation-metrics-section]{\textbf{Evaluation Metrics}}\\[2pt]
    \quad -- Threat score (primary), AUC, F1%
};

\node[colhdr, anchor=north west] (R2hdr) at ($(R1.south west)+(0,-0.3)$) {\hyperref[feature-importance-section]{(2) SHAP Feature Importance Evaluation}};

\node[secbox, anchor=north west] (R2) at ($(R2hdr.south west)+(0,-0.08)$) {%
    \textbf{SHAP Computation:}\\
    \quad -- Permutation SHAP (TabPFN); TreeSHAP (Random Forest, XGBoost, CatBoost)\\
    \quad -- Per fold: refit on train, explain held-out rows\\
    \textbf{Importance Evaluation:}\\
    \quad -- Beeswarm, fold-stability, cross-model ranking, feature-dependence plots\\
    \quad -- Basin-level SHAP importance maps%
};

\node[colhdr, anchor=north west] (R3hdr) at ($(R2.south west)+(0,-0.3)$) {(3) Synthetic Data Augmentation Evaluation};

\node[secbox, anchor=north west] (R3) at ($(R3hdr.south west)+(0,-0.08)$) {%
    \hyperref[synthetic-augment-section]{\textbf{Synthetic Data Generation:}}\\
    \quad -- Class-conditional TabPFN generator per fold\\
    \quad -- Fidelity (KS, Wasserstein-1, energy distance)\\
    \hyperref[augmentation-evaluation-section]{\textbf{Augmentation Evaluation:}}\\
    \quad -- All 15 models share identical synthetic data\\
    \quad -- Real rows duplicated then synthetics appended\\
    \quad -- Primary Metric: Per-model threat score $\Delta$%
};

\draw[dotted, line width=0.6pt]
    ($(Lhdr.north west)+(-0.18,0.18)$) rectangle ($(L5.south east)+(0.18,-0.18)$);
\draw[dotted, line width=0.6pt]
    ($(Rhdr.north west)+(-0.18,0.18)$) rectangle ($(R1.south east)+(0.18,-0.18)$);
\draw[dotted, line width=0.6pt]
    ($(R2hdr.north west)+(-0.18,0.18)$) rectangle ($(R2.south east)+(0.18,-0.18)$);
\draw[dotted, line width=0.6pt]
    ($(R3hdr.north west)+(-0.18,0.18)$) rectangle ($(R3.south east)+(0.18,-0.18)$);

\draw[arr] (L5.east) -- ++(0.4,0) |- (Rhdr.west);
\draw[arr] (R1.south) -- (R2hdr.north);
\draw[arr] (R2.south) -- (R3hdr.north);

\end{tikzpicture}
}
\caption{Conceptual overview of the experimental pipeline.
The left column lists the sequential steps applied within each outer training fold: data sources, features and response, cross-validation splitting, preprocessing, and hyperparameter tuning.
The right column shows the three analysis modules: the 15-model benchmark, SHAP feature importance evaluation, and TabPFN-based synthetic data augmentation evaluation.}
\label{fig:methodology}
\end{figure}

Figure~\ref{fig:methodology} summarizes the overall methodology.
The left column covers data compilation, features, cross-validation, preprocessing, and hyperparameter tuning.
The right column maps to the three primary contributions: the 15-model benchmark, SHAP feature importance, and TabPFN-driven synthetic data augmentation.
This section follows the same order.
We first describe the dataset and feature set, then the repeated stratified cross-validation design, within-fold preprocessing, and hyperparameter tuning.
We then present the model families, evaluation metrics, paired model comparisons, SHAP feature importance, and the TabPFN-based synthetic data augmentation procedure.

\subsection{Dataset}
\label{dataset-section}

\begin{table*}[t]
    \centering
    \caption{Features included in the model comparison.
    Features are grouped by category, with feature names, units, and brief descriptions.}
    \label{tab:used_features}
    \begin{tabularx}{\textwidth}{ll l X}
        \toprule
        \textbf{Category} & \textbf{Feature} & \textbf{Unit} & \textbf{Description} \\
        \midrule
        \textbf{Fire \& Terrain} & dNBR & - & Average differenced normalized burn ratio of watershed \\
         & PropHM23 & - & Proportion of watershed burned at high/mod severity, gradient $> 23^{\circ}$ \\
         & ContribArea & $km^2$ & Contributing area of observation location \\
        \midrule
        \textbf{Soil} & KF & - & Average KF-Factor (erodibility index of the fine fragments of the soil) of the watershed \\
        \midrule
        \textbf{Storm} & StormAccum & mm & Total rainfall accumulation of storm \\
         & StormDur & h & Total duration of storm \\
         & StormAvgI & mm/h & Average storm intensity \\
        \midrule
        \textbf{Rainfall} & PeakI15 & mm/h & Peak 15-minute rainfall intensity \\
         & PeakI30 & mm/h & Peak 30-minute rainfall intensity \\
         & PeakI60 & mm/h & Peak 60-minute rainfall intensity \\
        \bottomrule
    \end{tabularx}
\end{table*}

\newcolumntype{Y}{>{\Centering\arraybackslash}X}
\begin{table*}[]
\centering
\caption{Summary statistics of post-wildfire debris-flow predictors by response class.
For each feature, the table reports minimum, mean, median, maximum, and interquartile range (IQR).}
\label{tab:dataset_summary}
\begin{tabularx}{\textwidth}{l @{\hspace{1em}} YYYYY @{\hspace{1.5em}} YYYYY}
\toprule
\textbf{Feature} & \multicolumn{5}{c}{\textbf{No debris flow}} & \multicolumn{5}{c}{\textbf{Debris flow}} \\
\cmidrule(lr){2-6} \cmidrule(l){7-11}
 & Min & Mean & Med & Max & IQR & Min & Mean & Med & Max & IQR \\
\midrule
StormDur & 0.00 & 17.84 & 12.47 & 65.00 & 23.35 & 0.00 & 20.64 & 23.17 & 63.50 & 29.75 \\
StormAccum & 0.00 & 26.58 & 17.00 & 214.12 & 25.65 & 2.79 & 65.53 & 52.83 & 222.25 & 68.11 \\
StormAvgI & 0.00 & 3.79 & 1.40 & 58.67 & 2.65 & 0.00 & 8.84 & 3.20 & 58.67 & 4.12 \\
PeakI15 & 1.60 & 19.47 & 14.22 & 112.00 & 12.54 & 6.10 & 36.68 & 27.43 & 112.00 & 28.60 \\
PeakI30 & 1.20 & 13.92 & 10.00 & 80.00 & 9.20 & 3.56 & 26.38 & 19.60 & 80.00 & 18.17 \\
PeakI60 & 0.60 & 9.03 & 7.11 & 50.80 & 5.37 & 2.00 & 16.75 & 13.72 & 50.80 & 12.39 \\
ContribArea & 0.02 & 0.99 & 0.41 & 7.89 & 1.05 & 0.03 & 1.25 & 0.57 & 7.79 & 1.42 \\
PropHM23 & 0.00 & 0.46 & 0.49 & 0.99 & 0.49 & 0.01 & 0.56 & 0.59 & 0.99 & 0.30 \\
dNBR & 0.01 & 0.33 & 0.30 & 1.00 & 0.26 & 0.04 & 0.37 & 0.38 & 0.87 & 0.26 \\
KF & 0.00 & 0.21 & 0.24 & 0.98 & 0.11 & 0.00 & 0.30 & 0.24 & 11.36 & 0.06 \\
\bottomrule
\end{tabularx}
\end{table*}

\begin{figure}
    \centering
    \includegraphics[width=\linewidth]{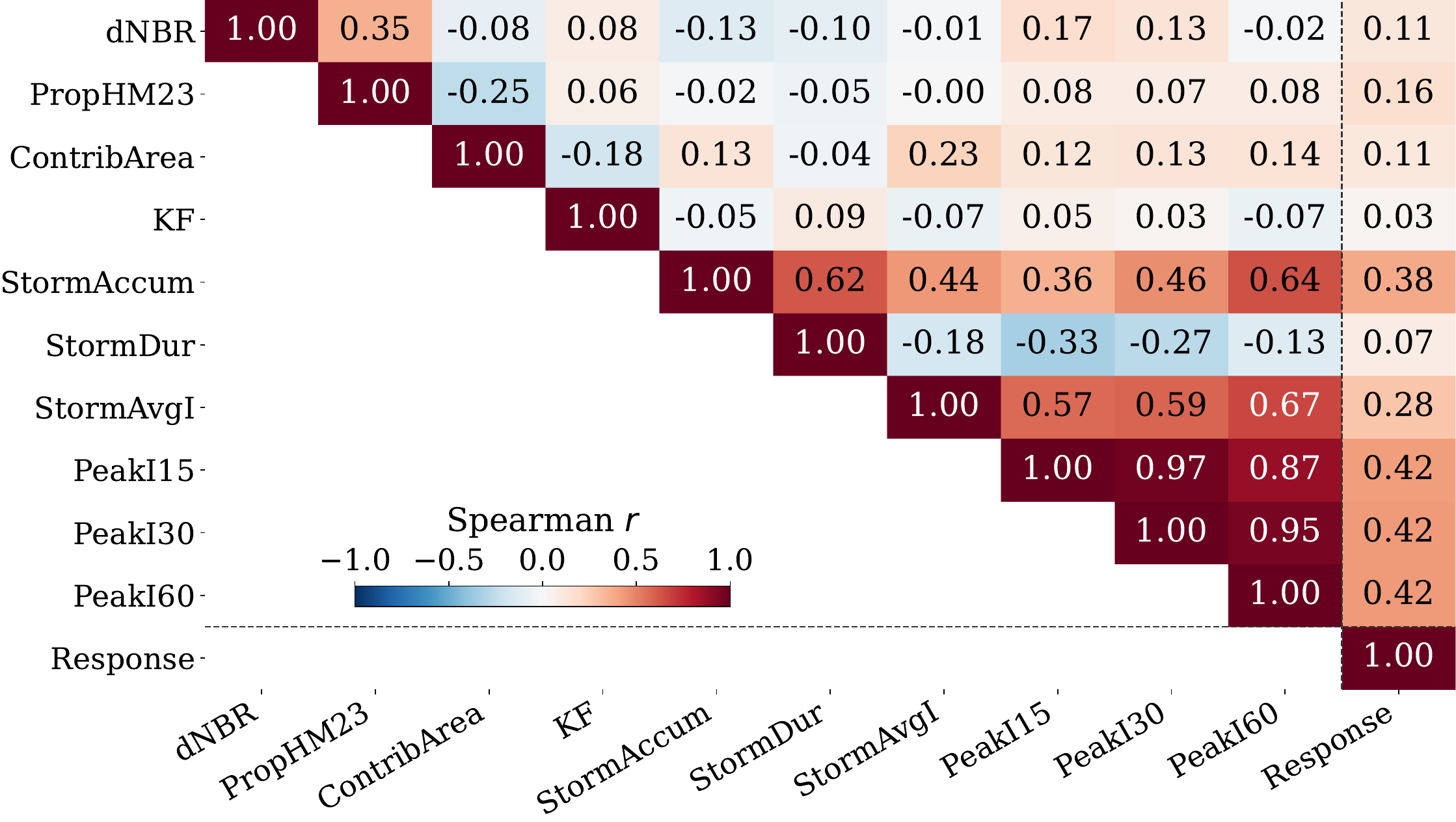}
    \caption{Upper-triangle Spearman rank correlation matrix for the ten benchmark features and the binary debris-flow response.
    Each cell gives a rank correlation value.
    Dashed lines separate the \textit{Response} row and column from the feature--feature entries.}
    \label{fig:feature_correlation}
\end{figure}

\begin{figure}
    \centering
    \includegraphics[width=\linewidth]{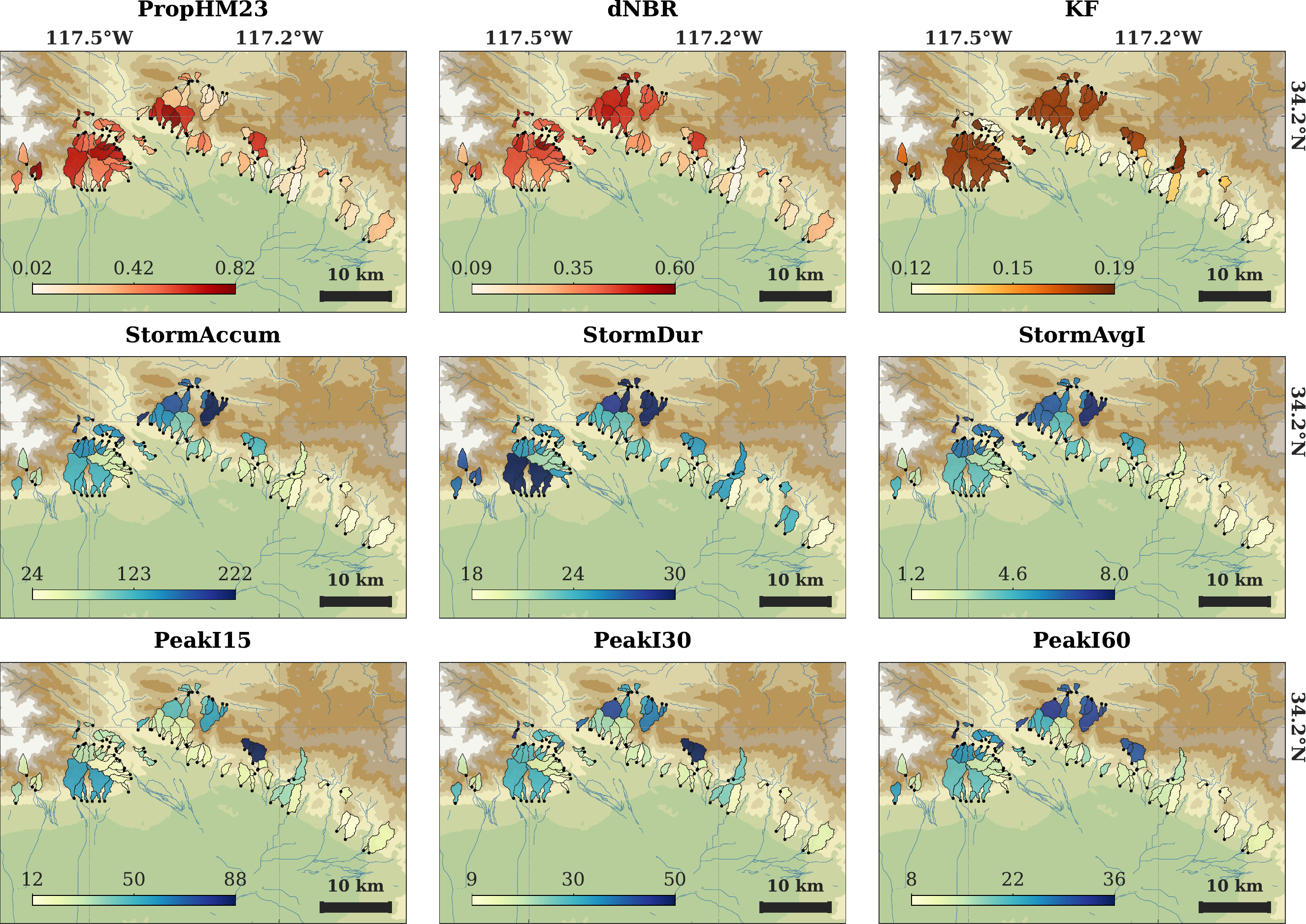}
    \caption{Grand Prix--Old fire: spatial distribution of nine features across delineated basins for a representative storm.
    The top row shows burn- and soil-related features, and the lower rows show rainfall accumulation, duration, mean intensity, and peak 15-, 30-, and 60-minute intensities.}
    \label{fig:feature_maps}
\end{figure}

\phantomsection
\label{dataset-data-sources}
For this evaluation, we used the publicly available basin-level dataset compiled by the USGS for post-wildfire debris-flow prediction \citep{staley_updated_2016}, which integrates response labels with rainfall, burn-severity, terrain, and soil features assembled under a consistent operational workflow.
The dataset contains 1,550 basin-storm observations from 34 fires and 244 storms collected between 2000 and 2013 across seven states in the western United States (California, Arizona, Colorado, Idaho, Montana, New Mexico, and Utah).
Each row corresponds to one burned basin under one storm event, and the response is binary: debris flow is positive, and no debris flow is negative, with 334 debris-flow and 1,216 no-debris-flow observations.
We chose this dataset because it underpins the operational USGS model \citep{staley_prediction_2017} and is a standard public benchmark for this prediction task, making our model rankings comparable to prior machine learning evaluations on the same problem and USGS data lineage \citep{kern_machine_2017, nikolopoulos_evaluation_2018, addison_assessment_2019, orland_scalable_2022, roten_machine_2022}.
The basin-level labels and features are also relatively high quality, having been compiled under a consistent USGS operational protocol, and the sample size and class imbalance are representative of post-wildfire debris-flow datasets more broadly, where confirmed events are scarce, and non-events dominate.
This makes findings on small-sample model behavior and class-imbalance handling more likely to transfer to other studies of this hazard.
The dataset is geographically concentrated, with 61\% of records from southern California fires and 39\% from the remaining states (Figure~\ref{fig:events_map}), reflecting the historical focus of the USGS operational debris-flow warning program on the western United States, where post-wildfire debris flows have frequently been observed.
To delineate the basin boundaries for visualization, we used the USGS postfire debris-flow Python package \citep{jonathan_m_king_pfdf_2025}.

\phantomsection
\label{dataset-features-response}
Table~\ref{tab:used_features} shows the ten numeric features used in the comparison, spanning rainfall, burn severity, basin properties, and soil erodibility.
Rainfall features are \textit{StormAccum}, \textit{StormDur}, \textit{StormAvgI}, \textit{PeakI15}, \textit{PeakI30}, and \textit{PeakI60} \citep{staley_updated_2016}.
Watershed and burn-severity features are \textit{ContribArea}, \textit{PropHM23}, and \textit{dNBR} \citep{key_landscape_2006}.
The soil feature \textit{KF} is the USLE soil erodibility factor ($K$) for the fine ($F$) fraction of the soil, derived from STATSGO \citep{schwarz_state_1995}.
The pairwise Spearman rank correlations among retained features and the response (Figure~\ref{fig:feature_correlation}) confirm that the three peak-intensity windows are strongly intercorrelated ($\rho > 0.90$) while burn-severity and soil features are largely independent of rainfall.
This characterizes the dependence structure of the feature space, with a strongly intercorrelated rainfall-intensity block alongside largely independent burn-severity and soil features, and establishes the marginal response associations against which the feature importance findings are later interpreted.

Figure~\ref{fig:feature_maps} complements this global view of feature correlations by mapping each feature across the basins of a single fire, showing the within-fire spatial structure that the global correlations average over.
Many of these features exhibit clear spatial autocorrelation within the fire, with neighboring basins carrying similar values.
This confirms that the features vary meaningfully at the basin level rather than being uniform within a fire, supporting the use of the basin as the modeling unit for this dataset.
The within-fire similarity among neighboring basins also reflects genuine physical structure in the data rather than label noise, and it informs the cross-validation design and the spatial interpretation of feature importances.

Three characteristics of the dataset make this a difficult prediction problem.
First, the two classes overlap heavily in feature space.
The per-feature summary statistics in Table~\ref{tab:dataset_summary} show that debris-flow and no-debris-flow basins differ in the expected directions on rainfall intensity, burn severity, and soil erodibility, but the distributions overlap substantially, and several features are nearly indistinguishable between the two classes.
The t-SNE projection of the feature space in Figure~\ref{fig:tsne_tabpfn_oof}a reinforces this, with the two classes overlapping heavily in the low-dimensional embedding and no clear cluster structure separating them.
Second, the dataset is class-imbalanced, with only 334 of 1,550 observations labeled as debris flow, a 21.5\% positive rate, so a model can reach high accuracy by defaulting to the majority no-debris-flow class while failing to identify the debris-flow events of interest.
Third, the dataset is small, with 1,550 observations across ten features, which limits the training data available to each model and increases the variance of the performance estimates.
Together, class overlap, class imbalance, and small sample size imply that no single feature or low-dimensional combination cleanly separates the classes, motivating the comparison of model families with differing inductive biases.

\subsection{Cross-Validation}
\label{cross-validation-section}
Given that the dataset is small and class-imbalanced, reliance on a single train-test split would yield unstable estimates that are sensitive to split choice.
For that reason, we evaluated the performance of the models with repeated stratified cross-validation \citep{kohavi_study_1995, arlot_survey_2010, krstajic_cross-validation_2014}.
In particular, we used stratified 5-fold cross-validation repeated 10 times, producing 50 outer evaluation splits with an equal percentage of no-debris-flow and debris-flow observations in each split.
Splits were precomputed once with a fixed random seed and reused unchanged across all models so that every model was evaluated on identical train-test splits.
Because 5-fold cross-validation was used, 20\% of observations were held out for testing, and 80\% were used for training.
All preprocessing and hyperparameter tuning were performed using only outer-training data.
Within each outer training fold, a stratified 80/20 split produced an inner training set used to fit candidate model configurations and a validation set used to score them during hyperparameter tuning, keeping the held-out test fold unseen during tuning to avoid selection bias \citep{cawley_over-fitting_2010}.
After hyperparameter selection, each model was refit on the full outer training fold using the chosen configuration so that the final fit used all available training data, while reported performance was measured on the held-out test fold that was never seen during tuning or fitting.

\subsection{Preprocessing of Features}
\label{preprocessing-section}
Preprocessing followed a consistent protocol intended to make comparisons across models methodologically fair.
Missing values were imputed within each outer training fold using a multi-step procedure that exploited known physical relationships among the features before falling back on summary statistics.
Missing entries among the 15-, 30-, and 60-minute peak rainfall intensity windows, and analogously the corresponding accumulation windows, were filled by fitting a log-log linear regression across durations, which corresponds to the empirical power-law intensity-duration relationship characteristic of rainfall \citep{koutsoyiannis_mathematical_1998}.
Decreasing peak intensity and increasing accumulation with longer durations were enforced for all rows.
Missing \textit{dNBR} values were imputed from \textit{PropHM23} using a fold-specific polynomial regression.
Any residual missing values were filled with the per-feature training median.

The imputer was fit on the outer training fold only and applied unchanged to the corresponding validation and held-out test splits to prevent information leakage.
Features were then provided to each model in the form most appropriate to its learning assumptions.
Distance-, kernel-, and gradient-based models received features standardized by a per-fold z-score transformation fit on the outer training data, while tree-based models, which are invariant to monotone feature transformations, operated directly on the features.
This per-model treatment of inputs was chosen to give each model its best chance of performing well on this dataset rather than to impose identical preprocessing across model families with different learning requirements.
Because the same imputation protocol was applied within each fold for every model, the same train-test splits were shared across the comparison, and scaling was applied only where required by the learning algorithm, observed performance differences can be attributed to model behavior rather than to differences in data handling.

\subsection{Hyperparameter Tuning}
\label{hyperparameter-tuning-section}
Hyperparameter selection followed a single, consistent protocol designed to keep model comparison fair across model families.
For each model and training condition, we used Bayesian optimization with 100 trials and Hyperband pruning to maximize validation threat score under fold-aware training \citep{akiba_optuna_2019, li_hyperband_2018}.
The selected configuration for each model was then fixed for final evaluation so that performance differences reflect model behavior rather than repeated retuning noise.
Tuning was conducted separately for unaugmented training and for augmented training with TabPFN-generated synthetic observations (Section~\ref{synthetic-augment-section}), so each configuration matched the training distribution under that condition.
Not all models were included in this shared tuning framework.
Some have no applicable hyperparameters under our setup, and others use their own native tuning procedures.
These differences are described alongside each model in the sections that follow.
As with the preprocessing protocol, these per-model differences reflect a deliberate choice to give each model its best chance of performing well rather than to enforce an artificially uniform tuning procedure across model families that would favor some models over others.

Two model-fitting choices that affect performance under class imbalance were held fixed rather than tuned per fold.
Class-weight rebalancing was applied through each library's standard balanced-weighting mode where supported, since this is the appropriate default for imbalanced binary classification \citep{he_learning_2009} and is not data-dependent in a way that would benefit from per-fold tuning.
Decision-threshold optimization was likewise not performed to avoid conflating hyperparameter selection with the metric used for model comparison.

\subsection{Models Compared}

\label{models-compared-section}
The 15 models span five model families, chosen so that the benchmark compares substantively different ways of learning from the same dataset rather than minor variants of one model architecture.

\hyperref[logistic-regression-section]{\emph{Logistic Regression Models}} are represented by two logistic regression variants: the operational Staley17 model \citep{staley_prediction_2017}, restricted to four features, and a standard logistic regression \citep{cox_regression_1958} trained on the full ten-feature set so that the link function can be isolated from feature-set restriction.

\hyperref[distance-kernel-section]{\emph{Distance- and kernel-based Models}} include K-Nearest Neighbors (KNN) \citep{cover_nearest_1967} and the Support Vector Classifier (SVC) \citep{cortes_support-vector_1995}.

\hyperref[deep-learning-section]{\emph{Deep-learning Models}} include a fully connected multi-layer perceptron (MLP) \citep{rumelhart_learning_1986} together with four models originally designed for sequential or image-like input: a one-dimensional convolutional network (CNN) \citep{lecun_gradient-based_1998}, a long short-term memory (LSTM) \citep{hochreiter_long_1997}, a Transformer encoder applied to per-feature token projections \citep{vaswani_attention_2017}, and a Mamba state-space model \citep{gu_mamba_2024}.

\hyperref[tree-based-section]{\emph{Tree-based Models}} include Random Forest \citep{breiman_random_2001}, Extremely Randomized Trees (ExtraTrees) \citep{geurts_extremely_2006}, eXtreme Gradient Boosting (XGBoost) \citep{chen_xgboost_2016}, Categorical Boosting (CatBoost) \citep{prokhorenkova_catboost_2018}, and tree-structured Recursive Feature Machines (xRFM) \citep{beaglehole_xrfm_2025}.

\hyperref[tabpfn-section]{\emph{Tabular Prior-Data Fitted Network (TabPFN)}} is in a family of its own; it represents a model class that performs prediction via in-context learning, pre-trained on millions of synthetic tasks \citep{hollmann_tabpfn_2023}.

\subsection{Logistic Regression Models}
\label{logistic-regression-section}
The \textbf{Staley17} \citep{staley_prediction_2017} logistic regression model employed in the USGS operational system defines the link function as:
\begin{equation}
    \chi = \beta + C_1TR + C_2FR + C_3SR
\end{equation}
where $C_1$, $C_2$, and $C_3$ are learned weights, $R$ denotes peak 15-minute rainfall accumulation in mm, $T$ is \textit{PropHM23}, $F$ is \textit{dNBR}, and $S$ is \textit{KF}.
This formulation ensures that when $R = 0$~mm, the predicted probability asymptotically approaches zero for sufficiently negative $\beta$ values.
Among the many candidate features examined by Staley et al., these four yielded the strongest predictive performance and were retained in the published model formulation \citep{staley_prediction_2017}.
Because the other 14 models consume \textit{PeakI15} directly as a 15-minute peak intensity, we derived the accumulation $R$ required by the Staley17 formulation as $R = \textit{PeakI15} \times 0.25$~h, which preserves the published equation while keeping all models conditioned on the same rainfall information.
In addition to Staley17, we included a standard \textbf{Logistic Regression} \citep{cox_regression_1958} trained on the full set of available features to isolate the effect of feature restriction from the choice of model family.
Staley17 was excluded from the shared hyperparameter tuning because it faithfully reproduces the original formulation.
Logistic Regression was included.

\subsection{Distance- and Kernel-based Models}
\label{distance-kernel-section}
Distance- and kernel-based models form class boundaries by finding local neighborhoods or maximum-margin separators in transformed feature space \citep{hastie_elements_2009}.
\textbf{K-Nearest Neighbors (KNN)} \citep{cover_nearest_1967} predicts class membership from the labels of nearby training observations under a chosen distance metric, with decision smoothness controlled by the neighbor count and distance weighting.
\textbf{Support Vector Classifier (SVC)} \citep{cortes_support-vector_1995} learns a separating hyperplane with maximum margin and, through nonlinear kernels, can represent curved class boundaries in the ten-dimensional feature space.
Both of these models used the shared hyperparameter tuning framework, with the tuned hyperparameters including neighborhood size and distance weighting for KNN and kernel, regularization, and kernel-scale settings for SVC.

\subsection{Deep-learning Models}
\label{deep-learning-section}
As a feed-forward deep-learning baseline, we included an \textbf{MLP} \citep{rumelhart_learning_1986} with fully connected layers and nonlinear activations, which imposes no structural prior over the input features.
We then evaluated four models whose inductive biases instead target sequential, spatial, or token-structured features: a one-dimensional \textbf{CNN} \citep{lecun_gradient-based_1998}, an \textbf{LSTM} \citep{hochreiter_long_1997}, a \textbf{Transformer} \citep{vaswani_attention_2017} encoder over feature tokens, and a \textbf{Mamba} \citep{gu_mamba_2024} state-space model.
These are not obvious choices for a small set of tabular, non-sequential features, but we included them to test whether convolution, recurrence, attention, or state-space structure can nevertheless extract useful signal from the feature ordering, a question that remains open in the tabular deep-learning literature \citep{grinsztajn_why_2022, borisov_deep_2024}.
All models in this family used the shared hyperparameter tuning framework, with tuned hyperparameters covering architectural width and depth. 
Optimizer and training-schedule parameters were held fixed across all deep-learning models.

\subsection{Tree-based Models}
\label{tree-based-section}
Tree-based models aggregate many decision trees to reduce variance and capture nonlinear interactions among features \citep{breiman_random_2001}, and tree-based approaches have demonstrated strong performance on post-wildfire debris-flow prediction \citep{roten_machine_2022}.
We included five tree-based models.
\textbf{Random Forest} \citep{breiman_random_2001} averages predictions across bootstrap-resampled trees with random feature subsampling at each split.
\textbf{Extremely Randomized Trees (ExtraTrees)} \citep{geurts_extremely_2006} extends this by selecting split thresholds at random, further reducing variance at a slight bias cost.
\textbf{eXtreme Gradient Boosting (XGBoost)} \citep{chen_xgboost_2016} and \textbf{Categorical Boosting (CatBoost)} \citep{prokhorenkova_catboost_2018} are gradient-boosted decision tree (GBDT) implementations that iteratively correct residual errors with shallow trees.
XGBoost applies regularized shrinkage and depth control while CatBoost uses ordered boosting to reduce overfitting.
\textbf{tree-structured Recursive Feature Machines (xRFM)} \citep{beaglehole_xrfm_2025} combines feature-learning kernel machines with an adaptive tree structure, allowing the kernel to adapt to local data geometry at each split of the input space.
All models except for xRFM used the shared hyperparameter tuning framework. 
xRFM used an internal tuning framework.

\subsection{Tabular Prior-Data Fitted Network (TabPFN)}
\label{tabpfn-section}

\begin{figure}
    \centering
    \includegraphics[width=\linewidth]{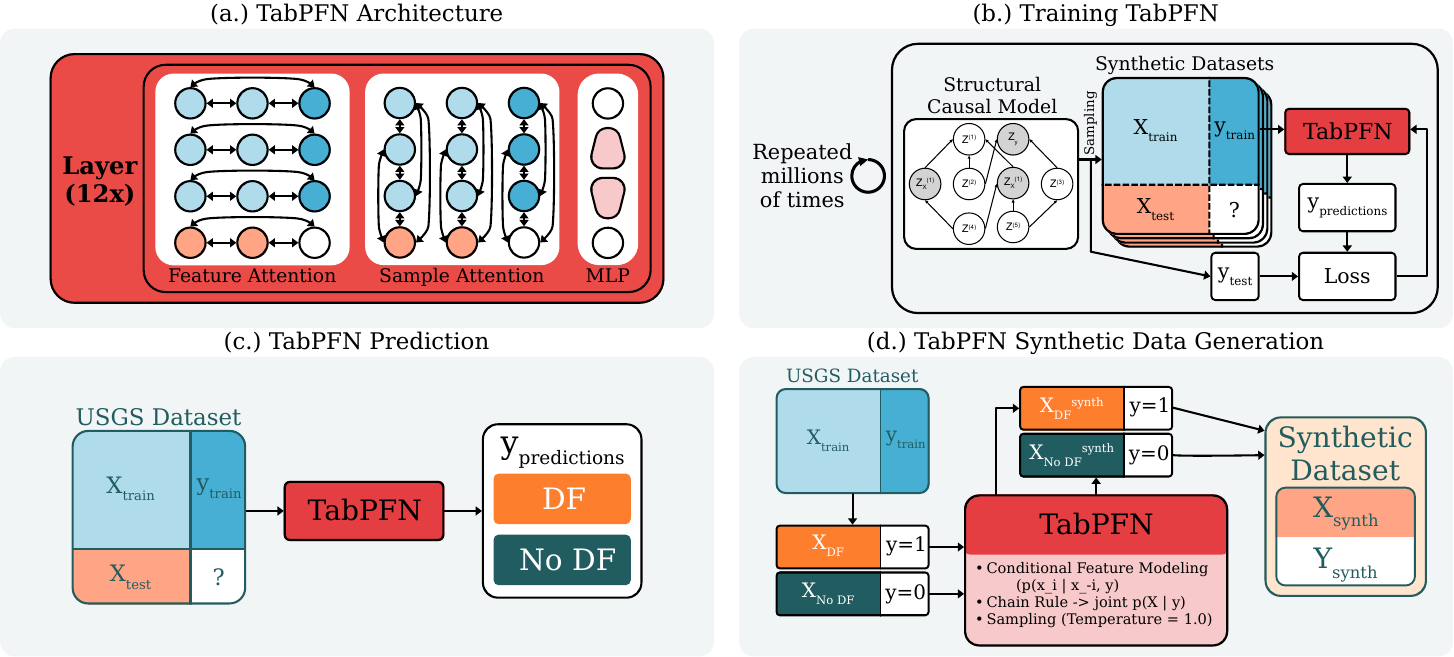}
    \caption{Overview of TabPFN architecture, training, prediction, and synthetic data generation.
    Panel~(a) shows the TabPFN layer block, repeated twelve times, combining feature attention, sample attention, and a multi-layer perceptron.
    Panel~(b) depicts TabPFN pre-training, where millions of synthetic datasets are sampled from a Structural Causal Model prior and used to train the Transformer weights via a supervised loss.
    Panel~(c) applies the pre-trained TabPFN to the USGS dataset: labeled training basins and unlabeled test basins enter the model together, and class probabilities (DF / No~DF) emerge in one forward pass with no parameter updates.
    Panel~(d) illustrates TabPFN-based synthetic data generation, in which class-conditional distributions are modeled autoregressively via conditional feature modeling ($p(x_i \mid x_{-i}, y)$), the chain rule yields the joint $p(X \mid y)$, and sampling produces synthetic debris-flow and no-debris-flow rows that augment the USGS training set.}
    \label{fig:tabpfn_diagram}
\end{figure}

Tabular Prior-Data Fitted Network (\textbf{TabPFN}) departs from the standard fit-from-scratch paradigm for machine learning.
Whereas conventional models such as gradient-boosted trees are trained anew on every dataset, TabPFN is a pre-trained Transformer that predicts on an unseen dataset in a single forward pass by treating the labeled training examples as context \citep{hollmann_tabpfn_2023}.
This capability is grounded in the Prior-Data Fitted Network framework, which reformulates supervised learning as a meta-learning task. 
The model is trained offline on millions of synthetic datasets sampled from a prior combining Structural Causal Models and Bayesian Neural Networks (Figure~\ref{fig:tabpfn_diagram}b), learning to approximate the posterior predictive distribution without any dataset-specific parameter updates \citep{muller_transformers_2024}.`
At inference time (Figure~\ref{fig:tabpfn_diagram}c), each training observation is encoded as a token, and the Transformer's self-attention layers jointly process the training context and the unlabeled test data to produce class probabilities, analogous to few-shot learning in large language models \citep{brown_language_2020} but specialized to tabular inputs.
We used TabPFN-2.5 \citep{grinsztajn_tabpfn-25_2025}, which extends the original architecture by incorporating large-scale open-source tabular data alongside the synthetic prior to improve generalization across diverse data distributions.

The internal architecture (Figure~\ref{fig:tabpfn_diagram}a) consists of twelve stacked layer blocks.
Each block combines feature-wise attention, which models dependencies among features, with sample-wise attention, which conditions predictions on the labeled training context, followed by a position-wise multi-layer perceptron.
This dual-attention design allows the model to simultaneously relate features to one another and relate test points to the labeled training distribution, a form of implicit non-parametric reasoning that requires no user-specified model of the data-generating process.
Three properties of this design are particularly relevant to the present comparison.
First, this deep architecture gives the model substantial capacity to learn complex, nonlinear interactions among features, well beyond the shallow functional forms of logistic regression.
Second, because the model is pre-trained on millions of synthetic tabular datasets, it brings transferable tabular inductive biases to small tabular datasets without requiring large domain-specific training data, a natural fit for the 1,550-observation USGS dataset.
Third, TabPFN requires no dataset-specific hyperparameter tuning, all model weights remain fixed after pre-training, and the model is applied with static settings, in contrast to the Bayesian search applied to the other tuned models.

Beyond prediction, the TabPFN software ecosystem includes an unsupervised extension that turns the same pretrained tabular Transformer into a class-conditional synthetic data generator (Figure~\ref{fig:tabpfn_diagram}d) \citep{grinsztajn_priorlabstabpfn-extensions_2026}.
Using this model as the synthetic data generator aligns with its architectural emphasis on modeling dependencies among tabular features and with offline meta-training across many synthetic tabular datasets, both of which support coherent conditional modeling when data is limited.

Thus, TabPFN serves two roles in this study: as one of the fifteen evaluated models, and as the generator of the synthetic training observations used for augmentation.

\subsection{Evaluation Metrics}
\label{evaluation-metrics-section}
Model performance was evaluated using a suite of metrics derived from the confusion matrix.
The dataset for this comparison is class-imbalanced, with no-debris-flow events outnumbering debris-flow events, so selecting metrics that remain informative under such skew is essential.
The four outcomes of binary classification are: true positives ($TP$, debris-flow events correctly predicted), false positives ($FP$, debris flow predicted but not observed), false negatives ($FN$, debris flow observed but not predicted), and true negatives ($TN$, non-events correctly identified).

The primary metric is the \textbf{threat score} (TS), also known as the Jaccard Score or Critical Success Index, defined as:
\begin{equation}
    TS = \frac{TP}{TP + FP + FN}
\end{equation}

In all metrics, debris flow is the positive class.
The threat score is particularly suited to hazard assessment because it disregards true negatives, which dominate imbalanced datasets, and measures how well the model recovers the event of interest relative to the total number of times that event was either predicted or observed \citep{schaefer_critical_1990}.
Supplementary metrics reported for completeness include accuracy, precision, recall, F1-score, and area under the receiver-operating-characteristic (ROC) curve (AUC).
Accuracy shows the proportion of correctly predicted events.
Precision and recall characterize the false-alarm and missed-event trade-off, respectively.
F1 is their harmonic mean.
ROC-AUC provides a threshold-independent measure of discrimination \citep{hanley_meaning_1982}.
For each metric, we reported the mean and standard deviation across all 50 outer test folds to characterize both central performance and fold-to-fold stability \citep{kohavi_study_1995, arlot_survey_2010}.
On a single inventory, fold-to-fold dispersion can be comparable to mean gaps among top models, so model comparisons must account for cross-validation variability rather than table rank alone \citep{krstajic_cross-validation_2014, nadeau_inference_2003}.

To assess whether differences between models are large relative to this variability, we compared pre-specified pairs using paired threat scores from the same outer splits, following single-dataset comparison guidance for models evaluated on identical resampled partitions \citep{dietterich_approximate_1998, nadeau_inference_2003}.
For each pair, we reported the mean paired difference $\Delta$TS, the fraction of folds in which one model achieved a higher threat score, and repeat-level summaries obtained by averaging threat score over the five folds within each of the ten CV repeats.
We summarized repeat-level $\Delta$TS with its mean, standard deviation, and 95\% confidence interval, and applied a two-sided Wilcoxon signed-rank test to the ten repeat-level differences as a nonparametric supportive check only \citep{demsar_statistical_2006}.
Because all folds draw from the same inventory, cross-validation scores are not independent observations, and repeat-level tests were not treated as full replication \citep{nadeau_inference_2003}.
Primary emphasis was on effect size, fold win rate, and overlap of repeat-level confidence intervals rather than on exhaustive pairwise hypothesis testing \citep{benavoli_time_2017}.

\subsection{SHAP Feature Importance Evaluation}
\label{feature-importance-section}
We used SHAP feature importances to interpret which features the different models rely on when predicting debris-flow probability on the held-out basins \citep{lundberg_unified_2017}.
SHAP assigns each feature a value representing its marginal contribution to the model output relative to a baseline, satisfying local-accuracy and consistency guarantees.
We chose SHAP over permutation-based importance because its per-observation additive scores support both global ranking and the basin-level spatial maps used here, and because Shapley values provide a principled additive decomposition relative to an explicit baseline.

Importance scores were computed on the same 50 outer test splits as the benchmark, so that SHAP estimates inherited the fold structure used for performance evaluation.
SHAP values are produced by an \emph{explainer}, an algorithm that estimates each feature's Shapley contribution for a given model class.
The choice of explainer depended on the model.
TabPFN used the library's built-in permutation explainer with a fixed evaluation budget \citep{grinsztajn_priorlabstabpfn-extensions_2026}.
Meanwhile, Random Forest, XGBoost, and CatBoost used TreeSHAP, an exact polynomial-time algorithm for tree-based models \citep{lundberg_explainable_2019}.
For each fold, the model was refit on the training split using the tuned hyperparameters from the main benchmark run, and SHAP values were then computed with the corresponding explainer on a uniformly subsampled subset of the held-out test rows.

For basin-level spatial analyses, we used a leave-one-fire-out procedure rather than the per-fold splits described above.
A TabPFN model was fit on all data except the target fire, and SHAP values were computed on the withheld basins using the same permutation explainer.
This ensured that all basins within a fire were explained by a single trained model, yielding spatially coherent importance polygons rather than a mosaic of per-fold explanations.
Basin-level maps included only basin observations from one storm event per fire, so rainfall-related features corresponded to a single meteorological event rather than a mixture of storms on the same terrain.
We focused this analysis on the Station and Grand Prix--Old fires because they have a large amount of observations for a single storm.

We then evaluated the SHAP feature importances along five complementary analyses.
First, we compared pooled-fold global rankings between TabPFN and the leading tree-based models (Random Forest, XGBoost, CatBoost) to identify points of agreement and divergence across model families.
Second, we examined the fold-to-fold stability of TabPFN's feature importance by tracking mean $|\mathrm{SHAP}|$ per feature across the 50 stratified outer folds.
Third, we mapped the resulting basin-level TabPFN SHAP values onto delineated watersheds, which addresses the limitation that global rankings cannot distinguish whether TabPFN conditions on basin-level feature variation within a fire.
Fourth, we related basin-level feature values to their SHAP contributions through Spearman rank correlation within each fire, which characterizes the dependence between feature value and importance independent of map geometry.
Fifth, we tested whether the sign of the SHAP--feature dependence aligns with established physical expectations by computing the Spearman correlation between each feature's values and its corresponding SHAP contributions pooled across all 50 folds for the four leading models, treating $|r|<0.15$ as a weak or negligible association.
We defined those expectations from previous post-wildfire debris-flow studies \citep{cannon_predicting_2010, staley_updated_2016, staley_prediction_2017} as a positive association between each feature and predicted debris-flow probability for all features except \textit{StormDur} and \textit{ContribArea}, for which no fixed directional match between classes was assumed.

Throughout this study, SHAP values are read as model-level feature reliance applicable to this study area and comparable burned-watershed settings, not as universal physical drivers of debris flows in all burned landscapes.

\subsection{Synthetic Data Generation}
\label{synthetic-augment-section}
Synthetic observations were generated using the TabPFN unsupervised extension \citep{grinsztajn_priorlabstabpfn-extensions_2026}.
Within each outer fold, a class-conditional generator was fit per class on that class's real training rows, and observations were drawn autoregressively one feature at a time from the learned conditional distributions, producing coherent multivariate observations without a fixed parametric form.

Generator fidelity was assessed per fold using the two-sample Kolmogorov--Smirnov statistic and Wasserstein-1 distance per feature, together with the multivariate energy distance between real and synthetic observations within each class \citep{massey_kolmogorov-smirnov_1951, villani_wasserstein_2009, szekely_energy_2013}.
These three metrics capture, respectively, the largest pointwise gap between univariate empirical distributions, the overall transport distance between those distributions, and any multivariate dependence mismatch across features.

\subsection{Synthetic Data Augmentation Evaluation}
\label{augmentation-evaluation-section}
Augmentation was treated as a fixed study component rather than a benchmark dimension, so a single synthetic generator was used throughout.
TabPFN's autoregressive conditional modeling was chosen because it is designed for small tabular datasets and is available within the same modeling framework.
The objective was to measure how this one consistent mechanism affects downstream model performance, not to compare generators.
Alternative tabular synthesis approaches, such as SMOTE-style interpolation \citep{chawla_smote_2002} and conditional generative adversarial networks \citep{xu_modeling_2019}, are therefore outside the scope of this study.

For each fold, synthetic observations were generated such that the count per class matched the real count of the opposing class, aiming for a balanced distribution while preserving the empirical per-class feature distributions.
However, to prevent synthetic data from overwhelming the training signal, the real observations were duplicated once before appending the synthetic rows, resulting in a final training set that was partially rebalanced rather than perfectly uniform.
Crucially, test sets remained entirely real, and all synthetic observations for a given fold were generated once and shared identically across all models, ensuring that benchmarking differences reflected model behavior rather than generator variance.

All 15 models were then compared again under the augmented training condition using the same outer cross-validation, inner tuning, and evaluation metrics as the unaugmented comparison, isolating the effect of augmentation on each model.

\section{Results}
%
We first compared models under unaugmented training.
Then we interpreted model behavior with SHAP-based feature importances.
Finally, we evaluated the quality of the synthetic data generation and the impact of augmentation on model performance.

\subsection{Model Comparison}

\begin{figure}
    \centering
    \includegraphics[width=0.5\linewidth]{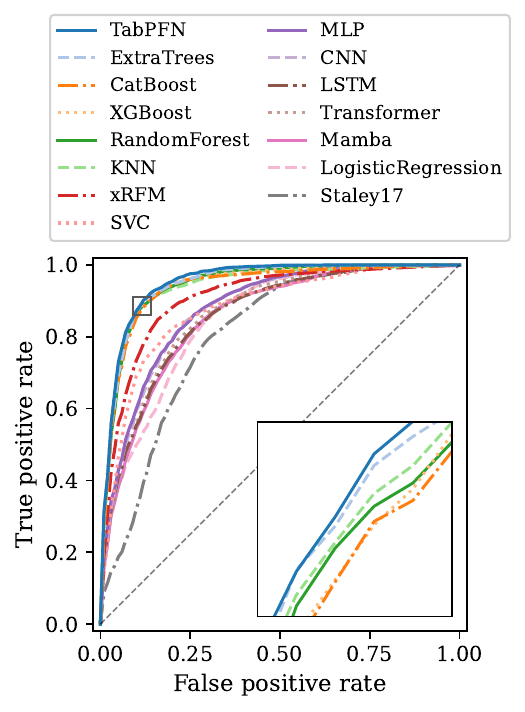}
    \caption{Mean ROC curves under unaugmented training, with one curve per model averaged over the same 50 stratified outer folds (10 repeats of 5 folds) and a dashed diagonal at random chance.
    The legend lists models in descending unaugmented threat score order, and the bottom-right inset magnifies the upper-left region.}
    \label{fig:roc_curves}
\end{figure}

\begin{figure}
    \centering
    \includegraphics[width=\linewidth]{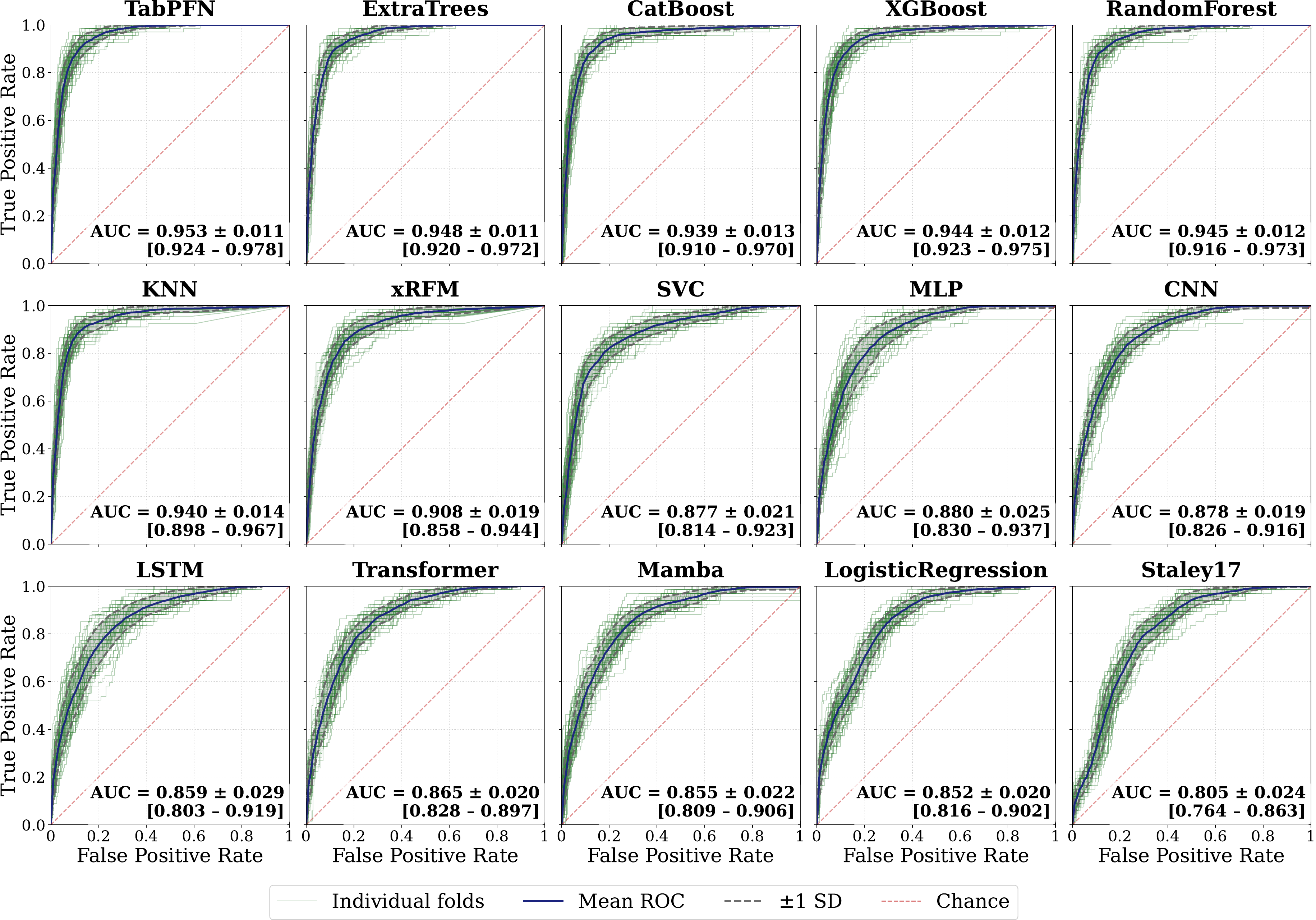}
    \caption{Per-model ROC traces under unaugmented training, arranged in a three-row by five-column grid sorted by descending mean AUC, with Staley17 in the bottom-right panel.
    Within each panel, thin green lines are the 50 individual outer-fold ROC curves, the navy line is the fold mean, the gray dashed lines are the $\pm 1$ standard deviation envelope, and the red dashed diagonal marks random chance.}
    \label{fig:detailed_roc_curves}
\end{figure}

\begin{table*}[ht]
\centering
\caption{Fifteen models under unaugmented training on the same 50 stratified outer folds.
For each metric, $\mu$ is the mean across folds and $\sigma$ is the standard deviation.
Rows are sorted by descending mean threat score (TS).
Within each metric, the three highest means are tinted by rank (1st: sky blue; 2nd: orange/gold; 3rd: bluish green).}
\label{tab:comparison}
\begin{tabularx}{\textwidth}{Xcccccccccc}
    \toprule
    Model & \multicolumn{2}{c}{Acc} & \multicolumn{2}{c}{F1} & \multicolumn{2}{c}{Recall} & \multicolumn{2}{c}{Prec} & \multicolumn{2}{c}{TS} \\
     & $\mu$ & $\sigma$ & $\mu$ & $\sigma$ & $\mu$ & $\sigma$ & $\mu$ & $\sigma$ & $\mu$ & $\sigma$ \\
    \midrule
    Staley17
      & 0.783 & 0.013
      & 0.262 & 0.049
      & 0.180 & 0.042
      & 0.504 & 0.092
      & 0.152 & 0.032 \\
    LogisticRegression
      & 0.773 & 0.026
      & 0.582 & 0.032
      & 0.732 & 0.046
      & 0.485 & 0.038
      & 0.412 & 0.032 \\
    Mamba
      & 0.766 & 0.030
      & 0.597 & 0.033
      & 0.801 & 0.061
      & 0.478 & 0.042
      & 0.426 & 0.034 \\
    Transformer
      & 0.769 & 0.033
      & 0.604 & 0.032
      & \cellcolor{cb3rd!40}{\textbf{0.811}} & 0.060
      & 0.484 & 0.044
      & 0.433 & 0.034 \\
    LSTM
      & 0.777 & 0.042
      & 0.606 & 0.051
      & 0.791 & 0.067
      & 0.495 & 0.060
      & 0.437 & 0.053 \\
    CNN
      & 0.789 & 0.024
      & 0.627 & 0.029
      & \cellcolor{cb2nd!40}{\textbf{0.822}} & 0.047
      & 0.508 & 0.034
      & 0.457 & 0.031 \\
    MLP
      & 0.794 & 0.032
      & 0.628 & 0.045
      & 0.802 & 0.063
      & 0.518 & 0.050
      & 0.459 & 0.048 \\
    SVC
      & 0.834 & 0.018
      & 0.666 & 0.032
      & 0.769 & 0.051
      & 0.589 & 0.036
      & 0.500 & 0.036 \\
    xRFM
      & 0.866 & 0.015
      & 0.673 & 0.042
      & 0.641 & 0.067
      & 0.714 & 0.050
      & 0.509 & 0.048 \\
    KNN
      & \cellcolor{cb2nd!40}{\textbf{0.899}} & 0.015
      & 0.755 & 0.039
      & 0.725 & 0.060
      & \cellcolor{cb1st!50}{\textbf{0.792}} & 0.045
      & 0.608 & 0.051 \\
    RandomForest
      & 0.898 & 0.018
      & 0.764 & 0.040
      & 0.769 & 0.057
      & \cellcolor{cb3rd!40}{\textbf{0.763}} & 0.054
      & 0.620 & 0.052 \\
    XGBoost
      & \cellcolor{cb3rd!40}{\textbf{0.898}} & 0.018
      & 0.765 & 0.041
      & 0.770 & 0.059
      & 0.763 & 0.051
      & 0.621 & 0.054 \\
    CatBoost
      & 0.894 & 0.016
      & \cellcolor{cb3rd!40}{\textbf{0.766}} & 0.034
      & 0.809 & 0.057
      & 0.730 & 0.042
      & \cellcolor{cb3rd!40}{\textbf{0.622}} & 0.045 \\
    ExtraTrees
      & 0.894 & 0.019
      & \cellcolor{cb2nd!40}{\textbf{0.777}} & 0.035
      & \cellcolor{cb1st!50}{\textbf{0.850}} & 0.046
      & 0.717 & 0.049
      & \cellcolor{cb2nd!40}{\textbf{0.636}} & 0.047 \\
    TabPFN
      & \cellcolor{cb1st!50}{\textbf{0.903}} & 0.016
      & \cellcolor{cb1st!50}{\textbf{0.777}} & 0.036
      & 0.784 & 0.057
      & \cellcolor{cb2nd!40}{\textbf{0.775}} & 0.050
      & \cellcolor{cb1st!50}{\textbf{0.637}} & 0.048 \\
    \bottomrule
\end{tabularx}
\end{table*}

\begin{figure}
    \centering
    \includegraphics[width=\linewidth]{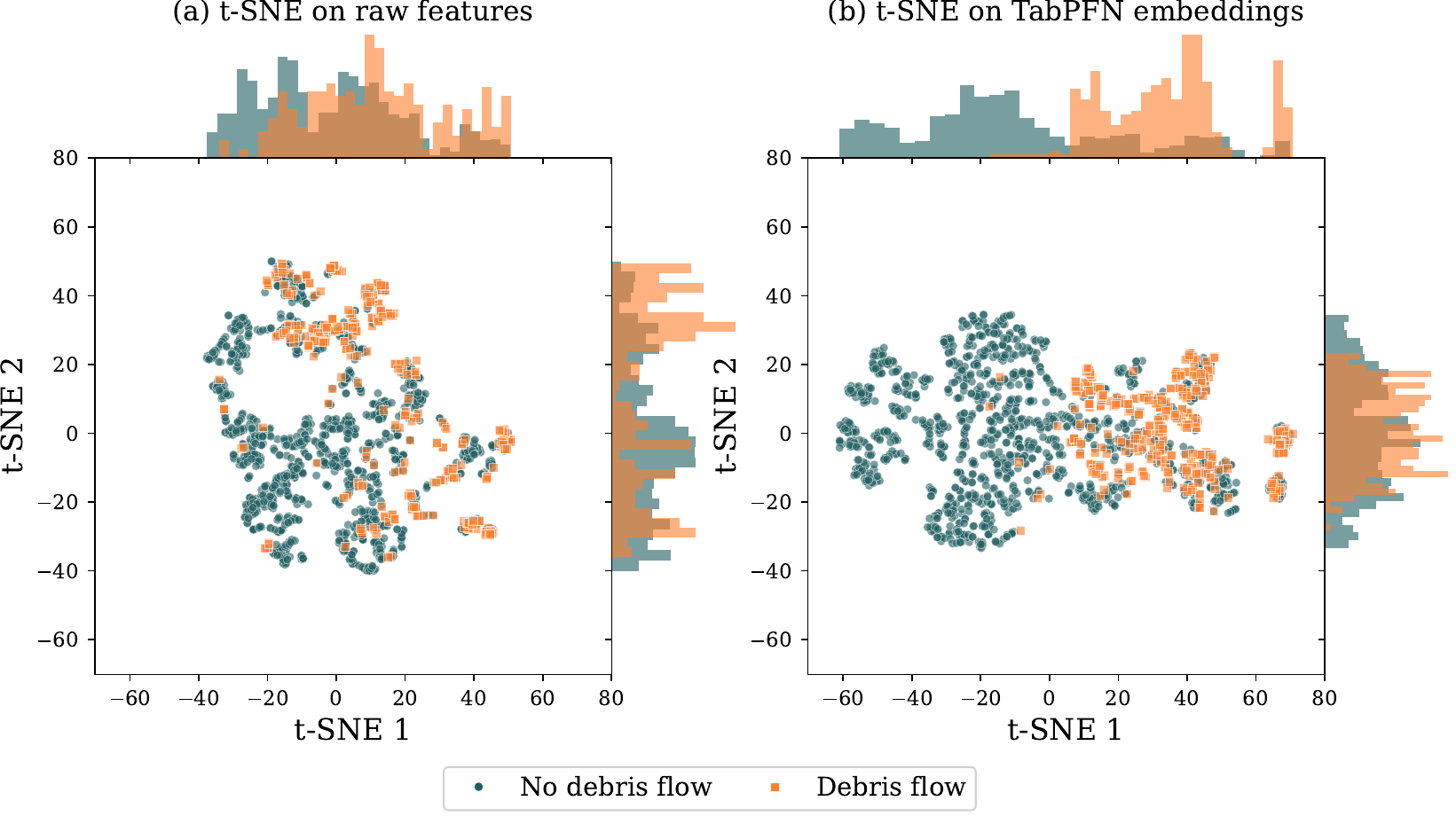}
    \caption{Two-panel t-SNE \citep{maaten_visualizing_2008} projection (perplexity 30) of the full dataset, with class-wise marginal histograms along each axis.
    Panel~(a) uses preprocessed raw features and panel~(b) uses out-of-fold TabPFN internal representations.}
    \label{fig:tsne_tabpfn_oof}
\end{figure}

Table~\ref{tab:comparison} summarizes the comparison under unaugmented training.
TabPFN ranked first with a mean threat score of 0.637, followed by a narrow top tier (span 0.017) comprising ExtraTrees (0.636), CatBoost (0.622), XGBoost (0.621), and Random Forest (0.620), with fold dispersion ($\sigma$) between 0.045 and 0.054 across these five models.
TabPFN and ExtraTrees tied for the highest F1 (both 0.777), and ExtraTrees achieved the highest recall (0.850).
Relative to this fold-to-fold variability, the leading models are closely matched.
The mean gap between TabPFN and ExtraTrees is only $\Delta$TS~$=+0.001$, TabPFN exceeded ExtraTrees on 30 of 50 paired folds, and the repeat-level 95\% confidence interval for $\Delta$TS ($-0.007$ to $+0.009$) includes zero.
Among the top five models, repeat-level rank~1 was split across ExtraTrees (five repeats), TabPFN (four), and CatBoost (one), so table order should not be read as a stable ordinal ranking within the top tier.
By contrast, TabPFN exceeded the operational Staley17 baseline by $\Delta$TS~$=+0.486$ on every fold with repeat-level 95\% CI $[+0.478,+0.493]$, and exceeded CatBoost by $\Delta$TS~$=+0.016$ (36 of 50 folds; repeat-level CI $[+0.005,+0.026]$).
Below the top tier, SVC (0.500) and xRFM (0.509) occupied a middle band, the four sequential or image-oriented deep-learning models clustered at 0.426--0.459 (Mamba, Transformer, LSTM, CNN) with high recall (0.79--0.82) but low precision (0.48--0.51), the ten-feature Logistic Regression reached 0.412, and the four-feature Staley17 model trailed at 0.152 with a recall of 0.180.

Figure~\ref{fig:roc_curves} shows the mean ROC curves averaged across the 50 outer folds for each model.
The threshold-independent ordering reproduces the threat score ranking: TabPFN (AUC 0.953), ExtraTrees (0.948), Random Forest (0.945), XGBoost (0.944), and CatBoost (0.939) trace the upper-left envelope, while the Staley17 model is relegated to a markedly lower curve at AUC 0.805.

Figure~\ref{fig:detailed_roc_curves} resolves the mean curves into their 50 per-fold traces and $\pm 1$~standard deviation envelopes.
TabPFN and the leading tree-based models occupy the most favorable ROC region across folds with tightly bundled traces, whereas the deep-learning and distance-based models exhibit visibly wider standard deviation envelopes.

\subsection{Feature Importance Evaluation}
\begin{table}[t]
\centering
\caption{SHAP-based feature rankings by model (1 = most important). Rankings from mean $|\mathrm{SHAP}|$ per feature. Feature numbers show overall rank averaged across models.}
\label{tab:shap_feature_rankings}
\setlength{\tabcolsep}{4pt}
\begin{tabularx}{0.51\columnwidth}{X *{4}{>{\centering\arraybackslash}c}}
\toprule
Feature & \makecell[c]{Random\\Forest} & XGBoost & CatBoost & TabPFN \\
\midrule
1.\ StormAccum & 2 & 2 & 1 & 2 \\
2.\ PeakI15 & 4 & 1 & 2 & 1 \\
3.\ PeakI30 & 1 & 6 & 8 & 3 \\
4.\ PropHM23 & 5 & 5 & 3 & 8 \\
5.\ PeakI60 & 3 & 10 & 5 & 5 \\
6.\ KF & 10 & 3 & 4 & 7 \\
7.\ StormAvgI & 7 & 4 & 9 & 6 \\
8.\ StormDur & 6 & 9 & 10 & 4 \\
9.\ dNBR & 8 & 8 & 7 & 9 \\
10.\ ContribArea & 9 & 7 & 6 & 10 \\
\bottomrule
\end{tabularx}
\end{table}

\begin{figure}
    \centering
    \includegraphics[width=1\linewidth]{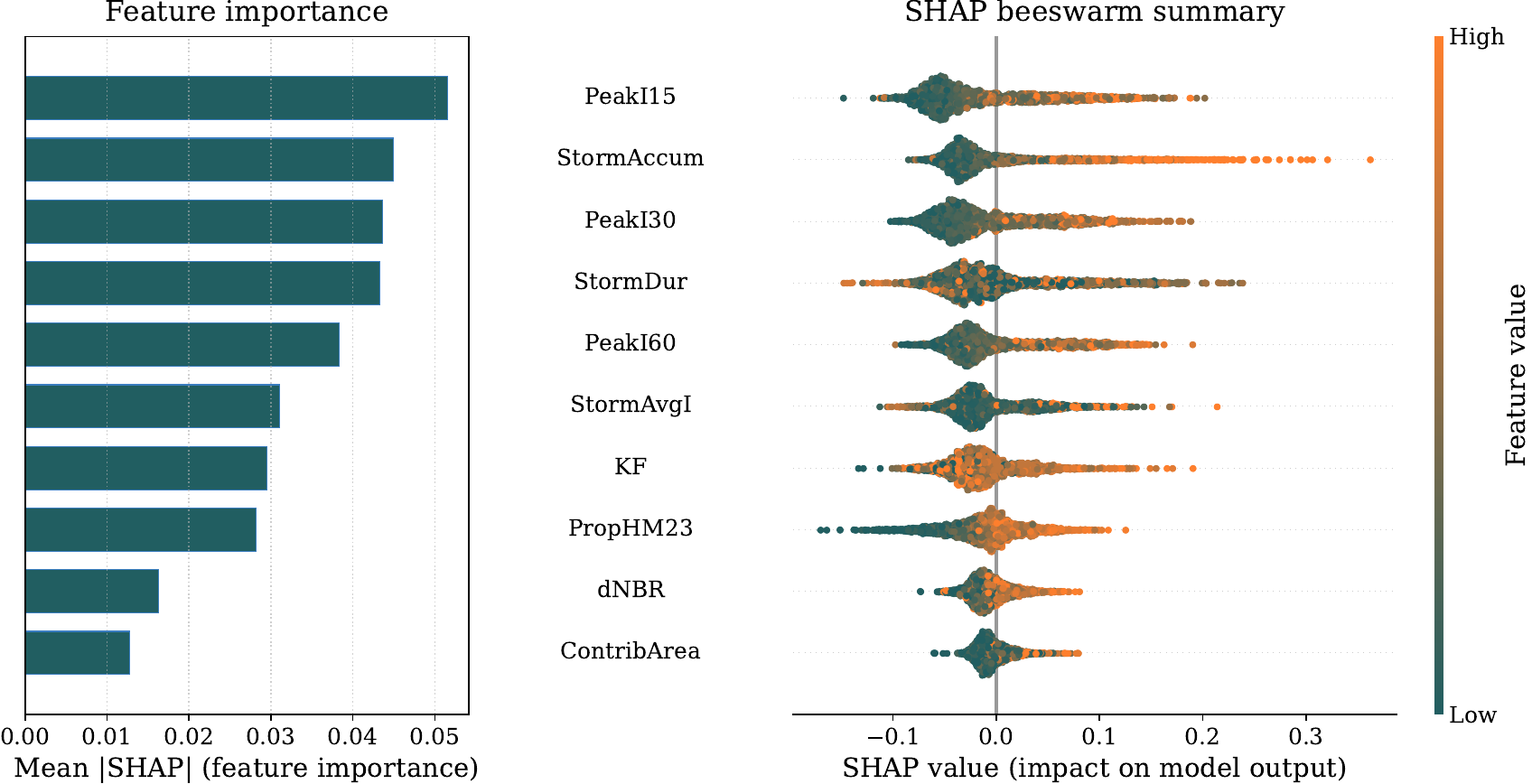}
    \caption{Combined SHAP summary for TabPFN, pooling held-out basin explanations across all 50 stratified outer folds.
    The left panel shows mean $|\mathrm{SHAP}|$ per feature as a horizontal bar chart, and the right panel shows the corresponding beeswarm in which each dot is one basin-level explanation plotted by its SHAP contribution to log-odds and colored by feature value from low to high.
    Features are ordered by mean $|\mathrm{SHAP}|$ so both panels share the same feature axis.}
    \label{fig:shap_summary}
\end{figure}

\begin{figure}
    \centering
    \includegraphics[width=\linewidth]{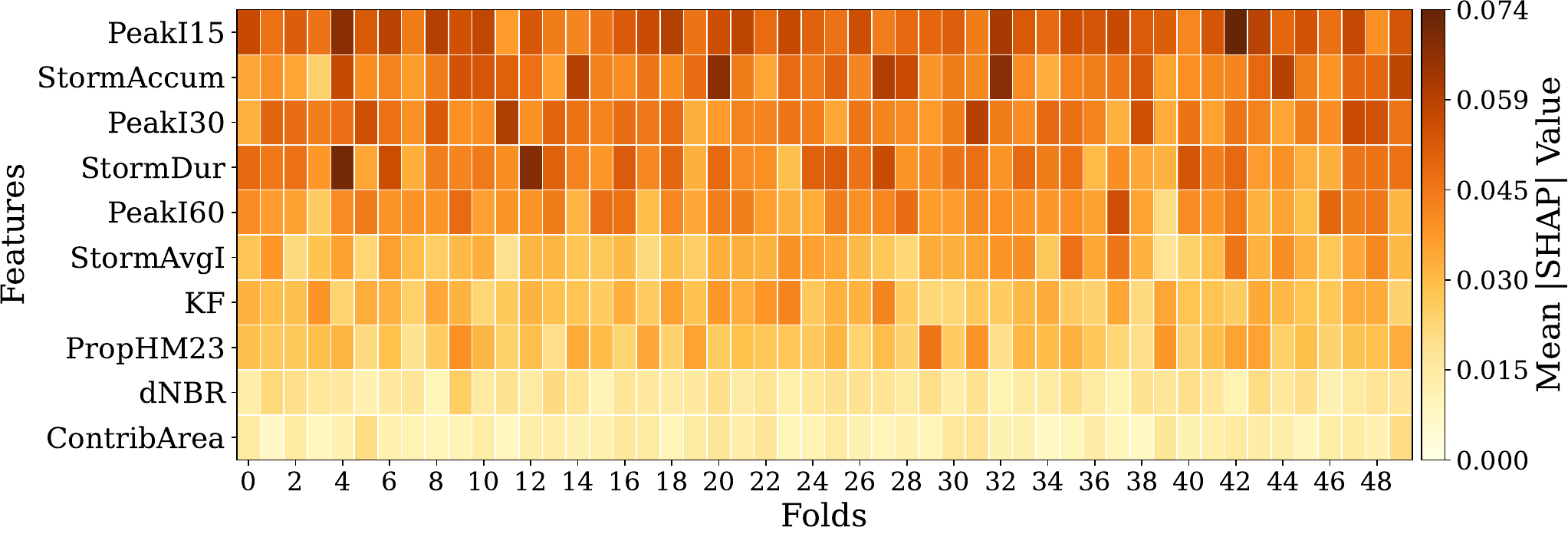}
    \caption{Fold-wise stability of TabPFN SHAP feature importance as a heatmap, with rows for the ten features and columns for the 50 stratified outer folds.
    Each cell gives the mean $|\mathrm{SHAP}|$ on the fold's held-out basins.}
    \label{fig:shap_stability}
\end{figure}

\begin{figure}
    \centering
    \includegraphics[width=0.6\linewidth]{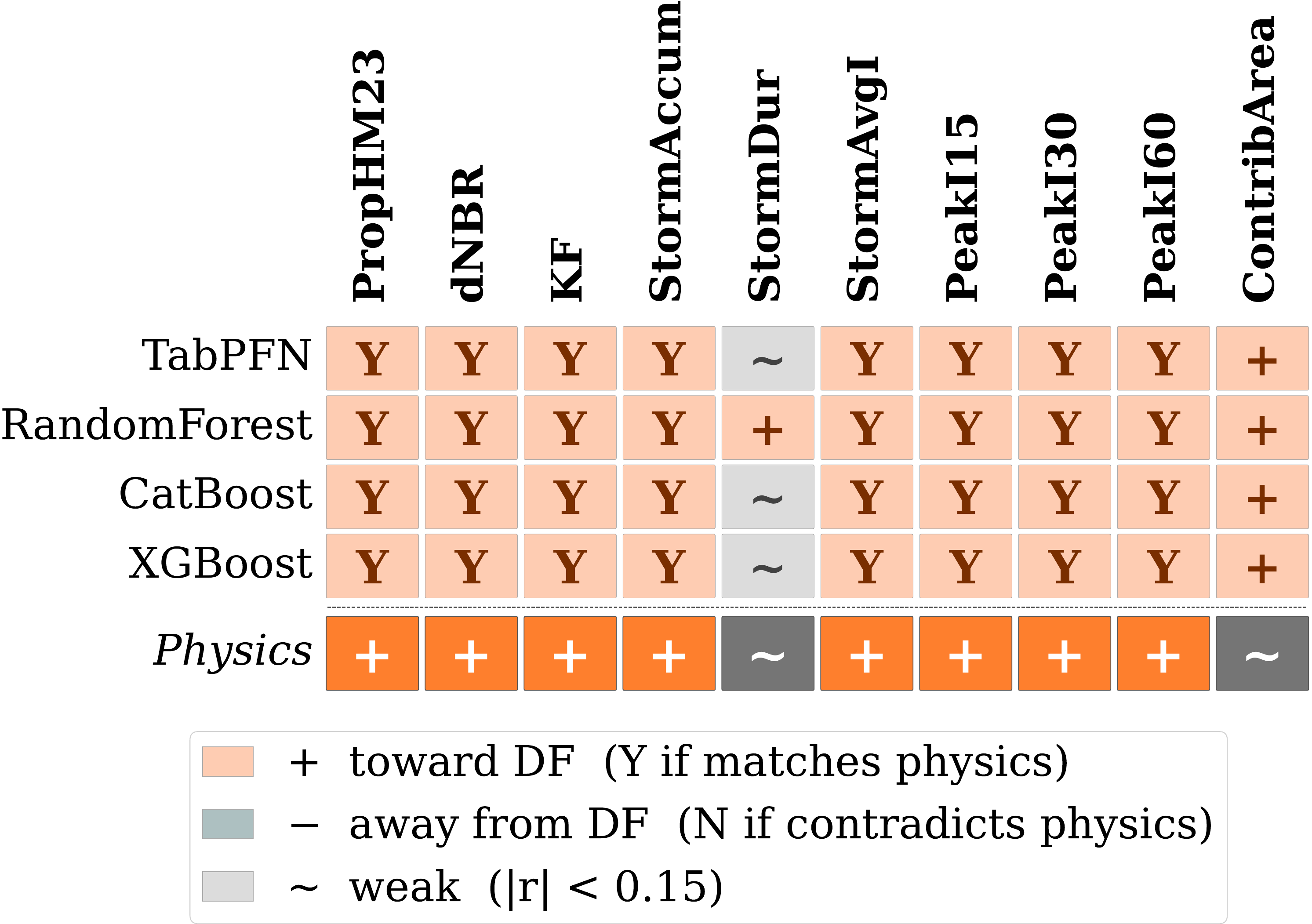}
    \caption{Agreement between SHAP-derived feature directions and physical expectations. Orange = positive direction (toward debris flow); teal = negative; gray = weak ($|r|<0.15$). Y/N indicates agreement or contradiction with physics; $+$/$-$ is shown directly for features with ambiguous physical sign. All four models correctly capture the expected positive direction for burn severity, terrain steepness, soil erodibility, and rainfall intensity.}
    \label{fig:shap_physics}
\end{figure}

The SHAP feature importance evaluation comprises four analyses: global rankings, fold-to-fold stability, basin-level spatial structure within fires, and alignment with physical expectations.
Rainfall intensity dominated global importance, while burn severity and soil erodibility showed the most consistent within-fire structure, and the four leading models agreed on physically expected directions.

The beeswarm summary (Figure~\ref{fig:shap_summary}) ranks peak 15-minute intensity, storm accumulation, and peak 30-minute intensity as the top three features by mean $|\mathrm{SHAP}|$, together carrying the bulk of the feature importance.
Table~\ref{tab:shap_feature_rankings} compares SHAP-derived rankings for TabPFN with high-performing tree-based models.
Rainfall metrics ranked highly across all four models, but burn severity and soil erodibility varied substantially.
Soil erodibility (\textit{KF}) ranked 3--4 for XGBoost and CatBoost but 10 for Random Forest, and \textit{PropHM23} ranked 3--5 for the tree-based models but 8 for TabPFN.

The fold-stability heatmap (Figure~\ref{fig:shap_stability}) shows that the three rainfall features (\textit{PeakI15}, \textit{StormAccum}, and \textit{PeakI30}) maintain high mean $|\mathrm{SHAP}|$ across nearly all folds, whereas \textit{dNBR} and \textit{ContribArea} exhibit lower and more variable importance from fold to fold.

Figures~\ref{fig:shap_spatial_stn} and~\ref{fig:shap_spatial_gpo} show SHAP maps and dependence scatters for the Station and Grand Prix--Old fires across four features (\textit{PeakI15}, \textit{PropHM23}, \textit{dNBR}, and \textit{KF}).
The spatial maps (top and middle rows) reveal contrasting behavior between the two fires.
On Grand Prix--Old, all four SHAP fields visually co-locate with their corresponding feature fields, with high-intensity, high-severity, and high-erodibility basins carrying positive SHAP.
On Station, only the \textit{dNBR} and \textit{KF} SHAP fields show the same visual co-location with their feature fields, while the \textit{PeakI15} and \textit{PropHM23} SHAP fields appear more uniform across basins than their feature fields.
The basin-level dependence scatters (bottom rows) quantify these patterns independent of map geometry.
On Grand Prix--Old, all four features show strong positive Spearman associations between feature value and SHAP contribution ($\rho = 0.67$--$0.92$).
On Station, \textit{dNBR} and \textit{KF} show similarly strong positive dependence ($\rho = 0.71$ and $0.75$), whereas \textit{PeakI15} and \textit{PropHM23} show only weak associations ($\rho = 0.27$--$0.45$).
Burn severity and soil erodibility therefore exhibit more consistent spatial structure and SHAP--feature coupling across both fires than peak 15-minute intensity and terrain steepness at these events.

Figure~\ref{fig:shap_physics} summarizes the SHAP--feature sign check for the four leading models against the consensus physical expectation.
All four correctly recovered a positive direction for burn severity, terrain steepness, soil erodibility, and every rainfall metric, with no contradictions observed for features with a clear physical sign.

\subsection{Synthetic Data Augmentation Evaluation}

\begin{table*}[ht]
\centering
\caption{Fifteen models under augmented training on the same 50 stratified outer folds and split assignments as Table~\ref{tab:comparison}.
For each metric, $\mu$ is the mean across folds with synthetic observations added during training, and $\Delta$ is the change relative to the unaugmented run.
Rows are sorted by descending mean threat score (TS).
Within each metric, the three largest improvements are tinted by rank (1st: sky blue; 2nd: orange/gold; 3rd: bluish green).}
\label{tab:comparison_augmented}
\begin{tabularx}{\textwidth}{Xcccccccccc}
    \toprule
    Model & \multicolumn{2}{c}{Acc} & \multicolumn{2}{c}{F1} & \multicolumn{2}{c}{Recall} & \multicolumn{2}{c}{Prec} & \multicolumn{2}{c}{TS} \\
     & $\mu$ & $\Delta$ & $\mu$ & $\Delta$ & $\mu$ & $\Delta$ & $\mu$ & $\Delta$ & $\mu$ & $\Delta$ \\
    \midrule
    Staley17
      & 0.772 & -0.011
      & 0.443 & \cellcolor{cb1st!50}{\textbf{+0.181}}
      & 0.421 & \cellcolor{cb1st!50}{\textbf{+0.241}}
      & 0.472 & -0.031
      & 0.285 & \cellcolor{cb1st!50}{\textbf{+0.133}} \\
    LogisticRegression
      & 0.773 & +0.000
      & 0.587 & +0.004
      & 0.745 & +0.013
      & 0.486 & +0.000
      & 0.416 & +0.004 \\
    CNN
      & 0.787 & -0.002
      & 0.624 & -0.003
      & 0.819 & -0.003
      & 0.506 & -0.003
      & 0.454 & -0.003 \\
    Mamba
      & 0.790 & \cellcolor{cb3rd!40}{\textbf{+0.025}}
      & 0.626 & +0.030
      & 0.812 & +0.011
      & 0.512 & \cellcolor{cb3rd!40}{\textbf{+0.033}}
      & 0.457 & +0.031 \\
    Transformer
      & 0.786 & +0.017
      & 0.627 & +0.023
      & 0.830 & +0.019
      & 0.506 & +0.021
      & 0.457 & +0.024 \\
    LSTM
      & 0.829 & \cellcolor{cb1st!50}{\textbf{+0.052}}
      & 0.677 & \cellcolor{cb3rd!40}{\textbf{+0.070}}
      & 0.826 & +0.034
      & 0.576 & \cellcolor{cb1st!50}{\textbf{+0.081}}
      & 0.513 & \cellcolor{cb3rd!40}{\textbf{+0.076}} \\
    SVC
      & 0.839 & +0.005
      & 0.688 & +0.022
      & 0.825 & +0.057
      & 0.592 & +0.003
      & 0.526 & +0.025 \\
    MLP
      & 0.841 & \cellcolor{cb2nd!40}{\textbf{+0.047}}
      & 0.698 & \cellcolor{cb2nd!40}{\textbf{+0.070}}
      & 0.853 & +0.051
      & 0.593 & \cellcolor{cb2nd!40}{\textbf{+0.075}}
      & 0.538 & \cellcolor{cb2nd!40}{\textbf{+0.078}} \\
    xRFM
      & 0.864 & -0.003
      & 0.722 & +0.049
      & 0.821 & \cellcolor{cb2nd!40}{\textbf{+0.180}}
      & 0.647 & -0.067
      & 0.566 & +0.058 \\
    KNN
      & 0.901 & +0.002
      & 0.774 & +0.019
      & 0.789 & \cellcolor{cb3rd!40}{\textbf{+0.064}}
      & 0.763 & -0.029
      & 0.632 & +0.024 \\
    ExtraTrees
      & 0.899 & +0.005
      & 0.778 & +0.002
      & 0.819 & -0.031
      & 0.744 & +0.026
      & 0.638 & +0.002 \\
    XGBoost
      & 0.899 & +0.001
      & 0.780 & +0.015
      & 0.826 & +0.056
      & 0.740 & -0.023
      & 0.640 & +0.019 \\
    RandomForest
      & 0.900 & +0.002
      & 0.780 & +0.016
      & 0.826 & +0.057
      & 0.742 & -0.022
      & 0.641 & +0.020 \\
    TabPFN
      & 0.897 & -0.006
      & 0.781 & +0.003
      & 0.846 & +0.062
      & 0.727 & -0.049
      & 0.641 & +0.004 \\
    CatBoost
      & 0.899 & +0.006
      & 0.782 & +0.016
      & 0.840 & +0.031
      & 0.734 & +0.004
      & 0.643 & +0.022 \\
    \bottomrule
\end{tabularx}
\end{table*}

\begin{figure}
    \centering
    \includegraphics[width=\linewidth]{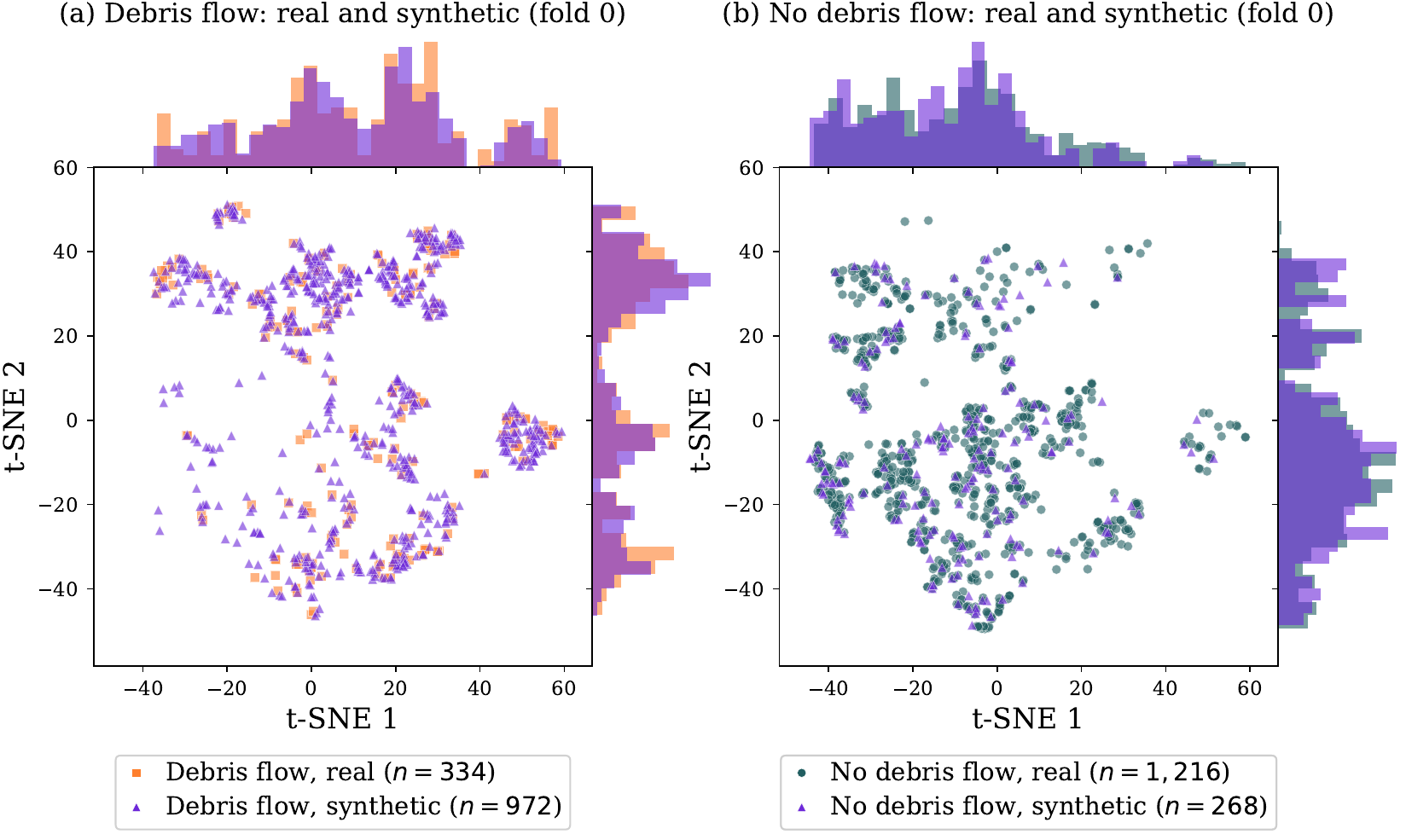}
    \caption{Two-panel t-SNE \citep{maaten_visualizing_2008} comparison of TabPFN-generated synthetic training rows against real rows for a single representative outer fold (fold~0), with class-wise marginal histograms along each axis.
    All four groups are projected under one global t-SNE fit (perplexity 30), so positions are comparable across panels.
    Panel~(a) shows real debris-flow basins and synthetic debris-flow observations, and panel~(b) shows real no-debris-flow basins and synthetic no-debris-flow observations.}
    \label{fig:synthetic_tsne}
\end{figure}

Before evaluating the impact of augmentation using synthetic observations, we assessed the realism of TabPFN-generated synthetic observations against the real training data.
Figure~\ref{fig:synthetic_distributions} shows class-conditional kernel density estimates for real and synthetic observations across the benchmark features, with per-feature Kolmogorov--Smirnov and Wasserstein summaries.
The synthetic-versus-real comparison indicates close agreement between classes.
The two-sample Kolmogorov--Smirnov $D$ ranges from 0.024 to 0.154 across features, with soil erodibility (\textit{KF}) the largest single-feature discrepancy and rainfall features mostly below $D = 0.10$, while the multivariate energy distance is 0.020 for no-debris-flow observations and 0.013 for debris-flow observations.

Figure~\ref{fig:synthetic_tsne} shows a two-panel t-SNE for one representative outer fold with class-wise marginal histograms, where panel~(a) compares real versus synthetic debris-flow observations and panel~(b) compares real versus synthetic no-debris-flow observations under a single t-SNE fit computed jointly across all four groups.
The t-SNE embedding confirms the agreement between synthetic and real observations visually, the synthetic points overlay the real point cloud in both classes without forming isolated sub-clusters.

Introducing the synthetic observations into training reshuffled the top tier within a narrow band (Table~\ref{tab:comparison_augmented}).
The five highest-threat-score models, CatBoost (0.643), TabPFN (0.641), Random Forest (0.641), XGBoost (0.640), and ExtraTrees (0.638), tightened into a span of only 0.005 in mean threat score, smaller than their typical fold $\sigma$ (0.040--0.042).
TabPFN exceeded ExtraTrees on 27 of 50 paired folds with a mean $\Delta$TS~$=+0.003$ and a repeat-level 95\% confidence interval that includes zero ($-0.006$ to $+0.012$), and CatBoost led TabPFN by $\Delta$TS~$=-0.003$ (25 of 50 folds in favor of TabPFN).
Repeat-level rank~1 among these five models was distributed across CatBoost (three repeats), ExtraTrees (three), Random Forest (two), and XGBoost (two), with no single model dominating.
CatBoost moved from third under unaugmented training to first under augmentation and also achieved the highest augmented F1 (0.782) among these top-performing models.

The largest gains appeared for the most restricted models.
Staley17 threat score rose by 13.3 percentage points and its recall by 24.1 percentage points, while the ten-feature Logistic Regression gained only 0.4 percentage points.
Among the deep-learning models, LSTM gained 7.6 percentage points in threat score and the MLP 7.8 percentage points, while CNN declined slightly by 0.3 percentage points.
Among middle-tier models, xRFM improved by 5.8 percentage points, SVC by 2.5 percentage points, and KNN by 2.4 percentage points.
The leading tree-based models and TabPFN improved by at most 2.2 percentage points in threat score.
Figure~\ref{fig:jaccard_delta} summarizes the per-model change in threat score.

\section{Discussion}

\subsection{Model Comparison}

TabPFN ranked first by threat score (0.637) and accuracy (0.903) but tied ExtraTrees on F1 (both 0.777), and ExtraTrees achieved the highest recall (0.850).
The mean threat-score margin between TabPFN and ExtraTrees ($\Delta$TS~$=+0.001$) is far smaller than fold-to-fold dispersion ($\sigma \approx 0.05$), and repeat-level comparisons do not support a durable ordering within the top tier, even though TabPFN clearly exceeds the Staley17 baseline and mid-tier models by large, stable margins.
The leading five nonlinear models, therefore, form a practical top tier rather than a single universally dominant model.
The mean ROC curves in Figure~\ref{fig:roc_curves} reproduce this ordering under a threshold-independent criterion, with TabPFN and the four leading tree-based models tracing the upper-left envelope and the Staley17 model relegated to a markedly lower curve, so the threat score ranking is not an artifact of the operating threshold.
This top-tier interpretation also explains why repeated cross-validation is necessary for model selection in this dataset \citep{kohavi_study_1995, krstajic_cross-validation_2014}.
Because the fold-level variation of each leading model exceeds the mean separation among those models, repeated evaluation supports a stable top-tier conclusion while discouraging over-interpretation of the exact within-tier order from any single split.
With top-tier performance this tightly grouped, the cost of obtaining it becomes the practical differentiator.
Tree-based models such as ExtraTrees, CatBoost, and XGBoost required Bayesian hyperparameter search, whereas TabPFN was run with static settings with no dataset-specific tuning and still matched or exceeded them.
For users with limited computational resources or rapid-response requirements, this tuning-free property could be as operationally relevant as the marginal threat score difference.

The tightness of the top tier is more consistent with a dataset-imposed ceiling than with the models being functionally equivalent.
The 1,550 basin-storm records and the southern California concentration of training observations jointly bound the achievable threat score regardless of model family.
The two-panel t-SNE in Figure~\ref{fig:tsne_tabpfn_oof} corroborates this directly: panel~(a) shows the two classes overlapping heavily in the raw feature space, while panel~(b) shows substantial separation under TabPFN's out-of-fold internal representations.
Even in panel~(b), residual overlap persists at the class boundary, indicating that TabPFN's learned embedding reduces but does not eliminate the class overlap in the available features.
Under these constraints, the tight clustering of top-tier threat scores reflects the limits of the available data as much as the capabilities of the models themselves.

The substantial gap between the logistic models and this top tier indicates that the feature-to-response mapping is not adequately represented by a fixed linear decision surface.
A linear model cannot express interaction terms or threshold effects among features, so both the four-feature Staley17 model (threat score 0.152) and the ten-feature Logistic Regression (0.412) trail the top nonlinear tier by 0.2 threat score or more, consistent with prior work showing that tree-based and neural models improve prediction over logistic regression \citep{addison_assessment_2019, roten_machine_2022}.
The intermediate value of the ten-feature Logistic Regression decomposes the Staley17-to-top-tier gap into two roughly equal contributions.
First, moving from Staley17 to the ten-feature Logistic Regression recovers about 0.26 threat score by relaxing the four-feature restriction alone, and second, moving from the ten-feature Logistic Regression to the top tier recovers a further 0.22 by relaxing the linear decision surface.
Feature breadth and decision-surface flexibility, therefore, both contribute to improved performance.

The four sequential or image-oriented deep-learning models cluster at threat score 0.43--0.46 with uniformly high recall (0.79--0.82) and low precision (0.48--0.51), indicating systematic over-prediction of debris-flow occurrence rather than training instability.
The per-fold ROC traces in Figure~\ref{fig:detailed_roc_curves} corroborate this reading.
The deep-learning panels show visibly wider $\pm 1$~SD envelopes than the top tree-based models and TabPFN, so their lower mean skill is accompanied by larger fold-to-fold variability rather than a tightly clustered but biased decision surface.
A plausible mechanism is a mismatch between model capacity and effective sample size.
Convolutional, recurrent, attentional, and state-space architectures are designed for image, sequence, or language domains and impose no inductive prior suited to tabular data, leaving them prone to overfitting on a dataset of this size.
This pattern is directionally consistent with tabular benchmarks, in which tree-based models often outperform high-capacity deep learning when no structure matched to tabular data is imposed \citep{grinsztajn_why_2022}.
The comparison is only indicative here because the inventory has roughly 1{,}550 observations and because the deep models apply convolutional, recurrent, attentional, and state-space architectures to a fixed feature ordering rather than standard tabular pipelines.

Several constraints should remain explicit, and they are directly connected to how well these operational considerations transfer beyond the fire settings represented in the inventory.
The dataset is geographically concentrated, with 61\% of records from southern California fires, which may favor features and model structures that generalize well within that region but perform differently in the drier, more continental environments represented by the remaining fires.
Because all observations are from the first two post-wildfire years, the models are not evaluated under later recovery conditions, when vegetation re-establishment and soil stabilization likely reduce debris-flow susceptibility \citep{degraff_timing_2015}.
Moderate class imbalance, with a positive-event rate of 21.5\%, is a further limitation.
Threat score remains informative at this rate, but operational deployment would still require recalibrating decision thresholds to the false-alarm and missed-event tolerances of the intended warning context.
Future work extending evaluation to multi-region datasets beyond the western United States and incorporating longer post-wildfire temporal windows would directly test whether these geographic and temporal limits affect transferability and robustness.

\subsection{Feature Importance Evaluation}

\begin{figure}
    \centering
    \includegraphics[width=\linewidth]{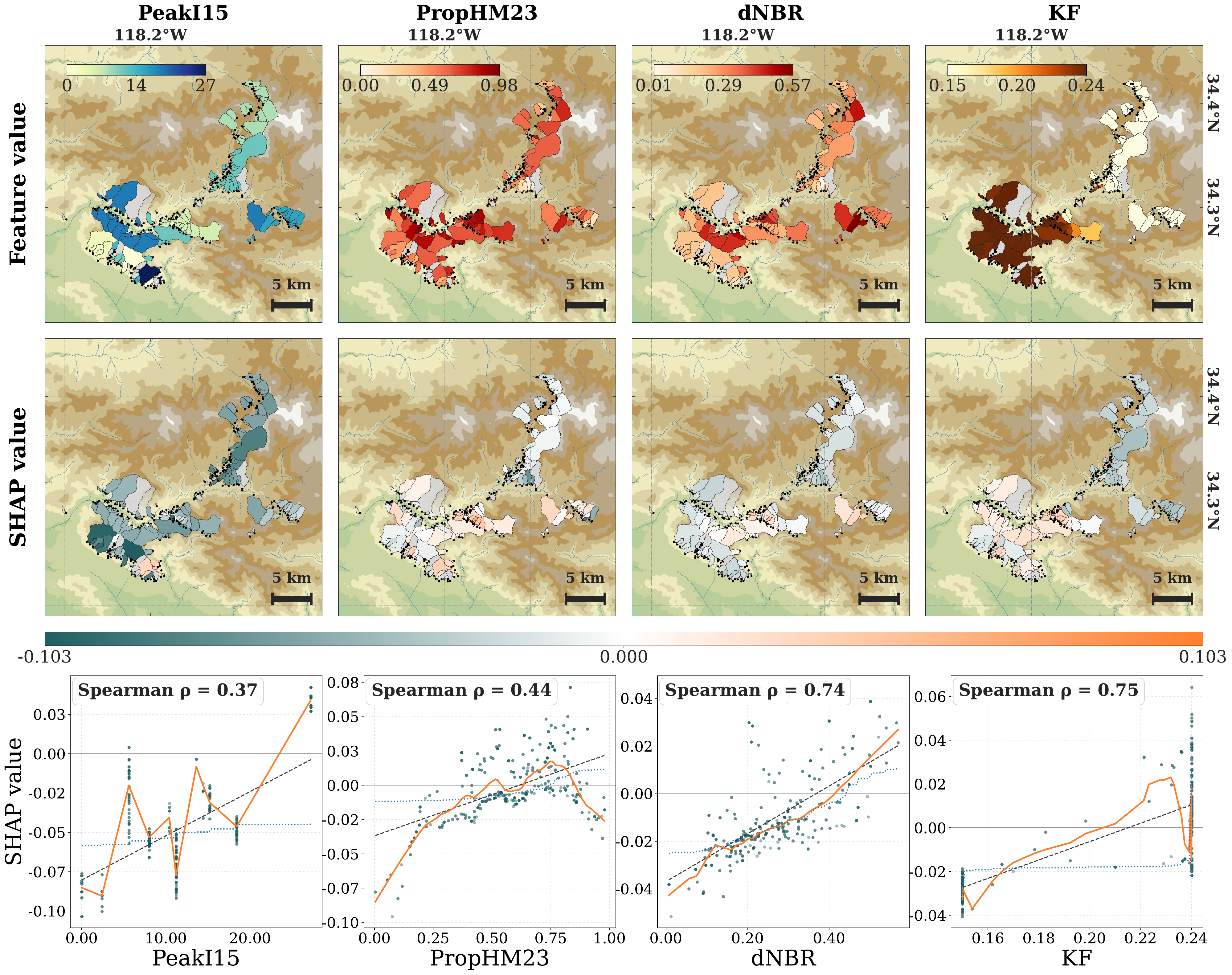}
    \caption{Basin-level SHAP feature importance for the Station fire in a three-row, four-column layout for \textit{PeakI15}, \textit{PropHM23}, \textit{dNBR}, and \textit{KF}.
    The top row shows feature-value maps and the middle row shows SHAP-value maps under a shared diverging colormap.
    The bottom row shows dependence scatter plots of SHAP value versus feature value, with teal dots for basins, a solid orange LOWESS curve, a dashed dark-gray OLS fit, a dotted blue rank-based fit, a horizontal gray reference at zero, and a text box reporting Spearman $\rho$.}
    \label{fig:shap_spatial_stn}
\end{figure}

\begin{figure}
    \centering
    \includegraphics[width=\linewidth]{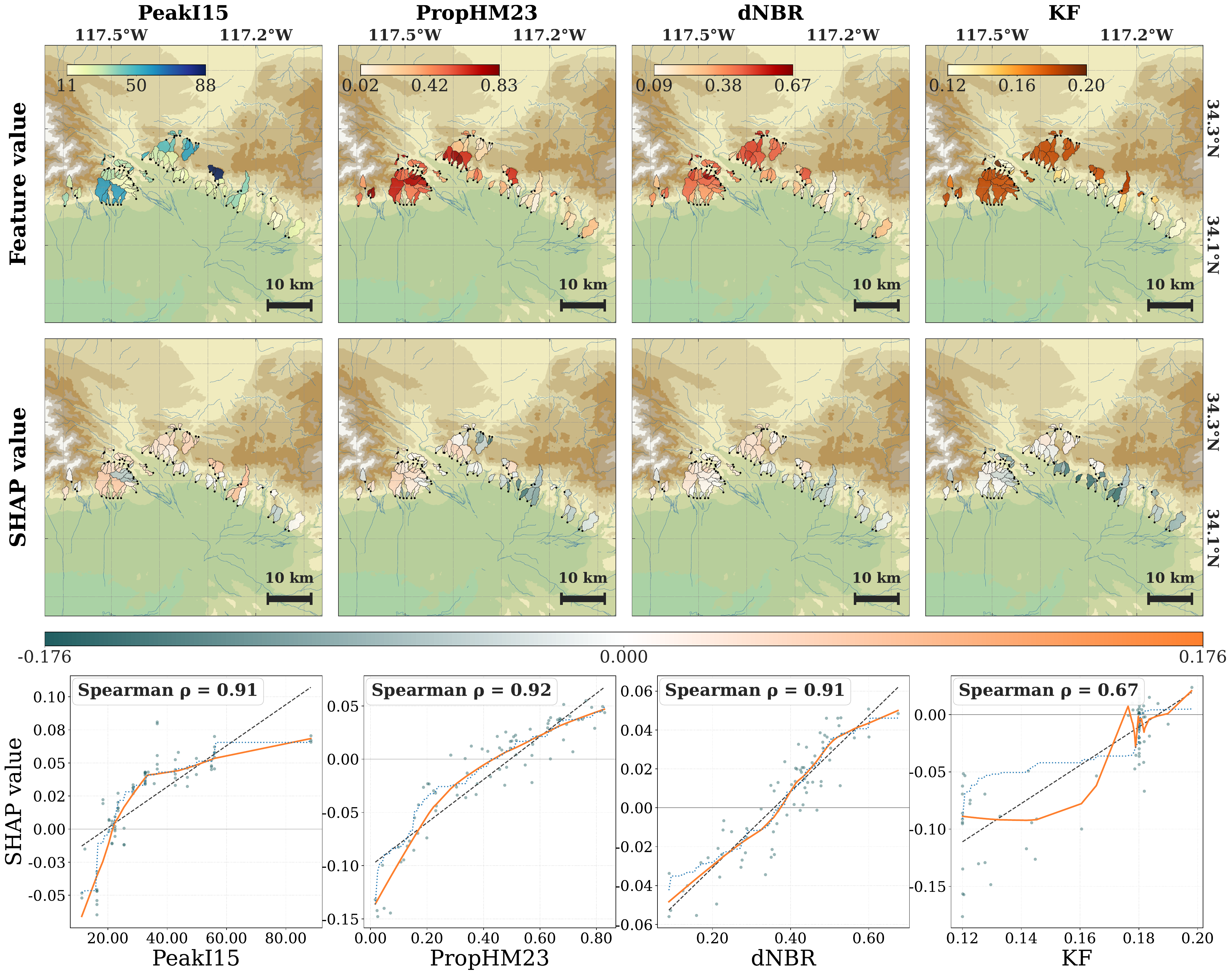}
    \caption{Basin-level SHAP feature importance for the Grand Prix--Old fire under the same three-row, four-column layout as Figure~\ref{fig:shap_spatial_stn} for \textit{PeakI15}, \textit{PropHM23}, \textit{dNBR}, and \textit{KF}.
    Row encoding follows Figure~\ref{fig:shap_spatial_stn}: feature-value maps on per-column sequential colormaps, SHAP-value maps on a shared diverging colormap, and dependence scatter plots with a solid orange LOWESS curve, a dashed dark-gray OLS line, a dotted blue rank-based line, a horizontal gray reference at zero, and a Spearman $\rho$ text box.}
    \label{fig:shap_spatial_gpo}
\end{figure}

SHAP analysis clarifies which features the evaluated models use when predicting debris-flow probability and supports interpretability for practitioners \citep{rundel_interpretable_2024, lundberg_unified_2017}.
For the four models we explained with SHAP, rainfall features occupy the top tier in every ranking (Table~\ref{tab:shap_feature_rankings}), indicating cross-model agreement on which features these models emphasize when predicting debris-flow probability.
Short-window intensity and storm accumulation receive the largest contributions, in line with the intensity--duration framing of the operational Staley17 model developed for the same regional warning context \citep{staley_prediction_2017, staley_updated_2016}.
Within the rainfall tier itself, the three peak-intensity windows are strongly intercorrelated ($\rho > 0.90$ in Figure~\ref{fig:feature_correlation}), so the specific rainfall feature each model promotes can shift while agreement among the four models on rainfall as the dominant predictive signal is preserved.
Below the rainfall tier, the rankings diverge substantially.
Burn severity (\textit{PropHM23}, \textit{dNBR}), soil erodibility (\textit{KF}), and contributing area receive different positions across the four models, indicating that agreement on the dominant predictive signal does not extend to which secondary feature carries residual predictive content.
Contributing area (\textit{ContribArea}) ranks consistently low across all four models, suggesting that basin size contributes little marginal signal once rainfall and burn-severity features are present.

The most striking divergences below the rainfall tier admit plausible mechanistic readings.
\textit{KF} ranks 3--4 for XGBoost and CatBoost but 10 for Random Forest.
Gradient-boosted trees can exploit a well-placed \textit{KF} threshold as a sequential corrective split once a rainfall feature has captured the dominant predictive signal, whereas Random Forest's split-level random feature subsampling means rainfall features are usually available at each split and \textit{KF} rarely surfaces as the chosen feature \citep{louppe_understanding_2013}.
\textit{PropHM23} ranks 3--5 for the tree-based models but 8 for TabPFN.
A plausible explanation is that TabPFN's in-context attention routes burn-severity signal through interaction terms with rainfall context that do not register cleanly in first-order marginal importance.
These divergent secondary-feature rankings across the four explained models reinforce that no single model's SHAP ranking should be read as a definitive statement of physical importance.

The four features Staley17 retained (\textit{PropHM23}, \textit{dNBR}, \textit{KF}, and 15-min rainfall accumulation) map only partially onto the SHAP rankings of the modern models.
Rainfall and \textit{KF} rank highly, but \textit{dNBR} ranks last or near-last across all four models, and \textit{PropHM23} is the lowest-ranked Staley17 feature in TabPFN (rank~8 of 10).
This is not a direct contradiction of the Staley17 model. 
It was selected to maximize logistic-regression skill under a four-feature constraint, and the features it retains may exploit different structural niches than those on which modern nonlinear models most rely.
The constructive reading is that rainfall, \textit{KF}, and \textit{PropHM23} receive consistently high SHAP ranks across the four explained models, while \textit{dNBR} contributes little marginal SHAP signal once those features are present.

TabPFN's mean absolute SHAP values (Figure~\ref{fig:shap_summary}, left panel) place short-window peak intensity (15~min), storm accumulation, and 30-minute peak intensity as the three largest contributors to predicted debris-flow probability, followed by storm duration and 60-minute peak intensity.
The beeswarm panel of Figure~\ref{fig:shap_summary} resolves the sign and per-basin spread behind these magnitudes.
For the top rainfall features, high values sit on the positive side and low values on the negative side, confirming a monotonic mapping from rainfall intensity and accumulation to predicted debris-flow probability at the basin level rather than an averaging artifact.
Lower-ranked features cluster more tightly near zero with mixed color, indicating that their small mean $|\mathrm{SHAP}|$ reflects genuinely weak per-observation effects rather than cancellation between large positive and negative contributions.

The fold-stability heatmap (Figure~\ref{fig:shap_stability}) shows that rainfall features maintain high mean $|\mathrm{SHAP}|$ across nearly all 50 outer folds, whereas burn ratio (\textit{dNBR}) and contributing area exhibit lower and more variable importance.
This asymmetry indicates that the rainfall-driven importances are reproducible across training folds, while the burn-severity and terrain importances are more sensitive to which folds the model is fit on.
The rainfall ranks can therefore be read as a stable property of TabPFN feature importance, while the \textit{dNBR} and contributing-area ranks are less trustworthy as point estimates and are better reported with fold-level uncertainty.
This within-model instability also lines up with the cross-model picture in Table~\ref{tab:shap_feature_rankings}.
The features whose TabPFN importance is least fold-stable are the same features whose rank disagrees most across CatBoost, Random Forest, and XGBoost, so low-signal features are unstable both within TabPFN across folds and across model families, while rainfall is stable on both axes.

The basin-level maps in Figures~\ref{fig:shap_spatial_stn} and~\ref{fig:shap_spatial_gpo} let us read the spatial structure of the SHAP field directly against the spatial structure of each feature field.
Because the features themselves are spatially autocorrelated within a fire (Section~\ref{dataset-section}), this comparison is a non-trivial test of whether the model is conditioning on the feature rather than on a global bias.
If the model were ignoring a feature, the SHAP map for that feature would be spatially flat or unrelated to the feature map.
Instead, on the Grand Prix--Old fire each of the four Staley17-style features produces a SHAP map whose high and low patches track the corresponding feature map (Figure~\ref{fig:feature_maps}), with basins of high storm intensity, high \textit{PropHM23}, high \textit{dNBR}, and high \textit{KF} carrying positive SHAP and the converse basins carrying negative SHAP.
On Station, \textit{dNBR} and \textit{KF} SHAP fields mirror the spatial variation of those feature fields, while the SHAP maps for peak 15-minute intensity and \textit{PropHM23} show weaker spatial correspondence.
This co-variation between the feature map and the SHAP map is the spatial signature of the model conditioning its prediction on the physical quantity rather than on a global bias or on incidental basin identity.

The dependence scatter plots in the bottom row of each figure make this relationship direct.
SHAP value is plotted against feature value, so a monotone trend indicates the model is mapping more of the feature into a larger contribution to debris-flow probability.
On Grand Prix--Old all four features show clear monotonic SHAP--feature dependence, with positive Spearman $\rho$ for the rainfall, burn-severity, and erodibility features, indicating that the model has internalized the expected sign for each feature with a clear physical direction.
On Station, \textit{dNBR} and \textit{KF} retain strong monotonic dependence ($\rho = 0.71$ and $0.75$), while \textit{PeakI15} and \textit{PropHM23} flatten ($\rho = 0.27$--$0.45$), showing that the contribution of those two features to the prediction varies little with their basin-level values at this event.
Read together for these two fires, the agreement between the spatial structure of the feature and SHAP maps and the monotonic SHAP--feature scatter relationships indicates that TabPFN's predictions are guided by feature--response relationships with the expected sign rather than by spurious associations.

Beyond feature reliance, a meaningful concern for any data-driven hazard prediction model is whether the learned relationships point in the physically expected direction.
Figure~\ref{fig:shap_physics} addresses this directly.
All four models assign positive SHAP contributions to higher values of burn severity, terrain steepness, soil erodibility, and every rainfall metric, with no reversals observed for any feature with a clear physical sign.
This concordance between learned and expected directions supports treating the top-tier models not merely as pattern-matching devices but as learning feature--response relationships with the expected sign for rainfall, burn severity, and soil erodibility, strengthening the case for their use alongside the operational Staley17 framework in the regional warning context this dataset was assembled for.

\subsection{Synthetic Data Augmentation Evaluation}

\begin{figure}
    \centering
    \includegraphics[width=\linewidth]{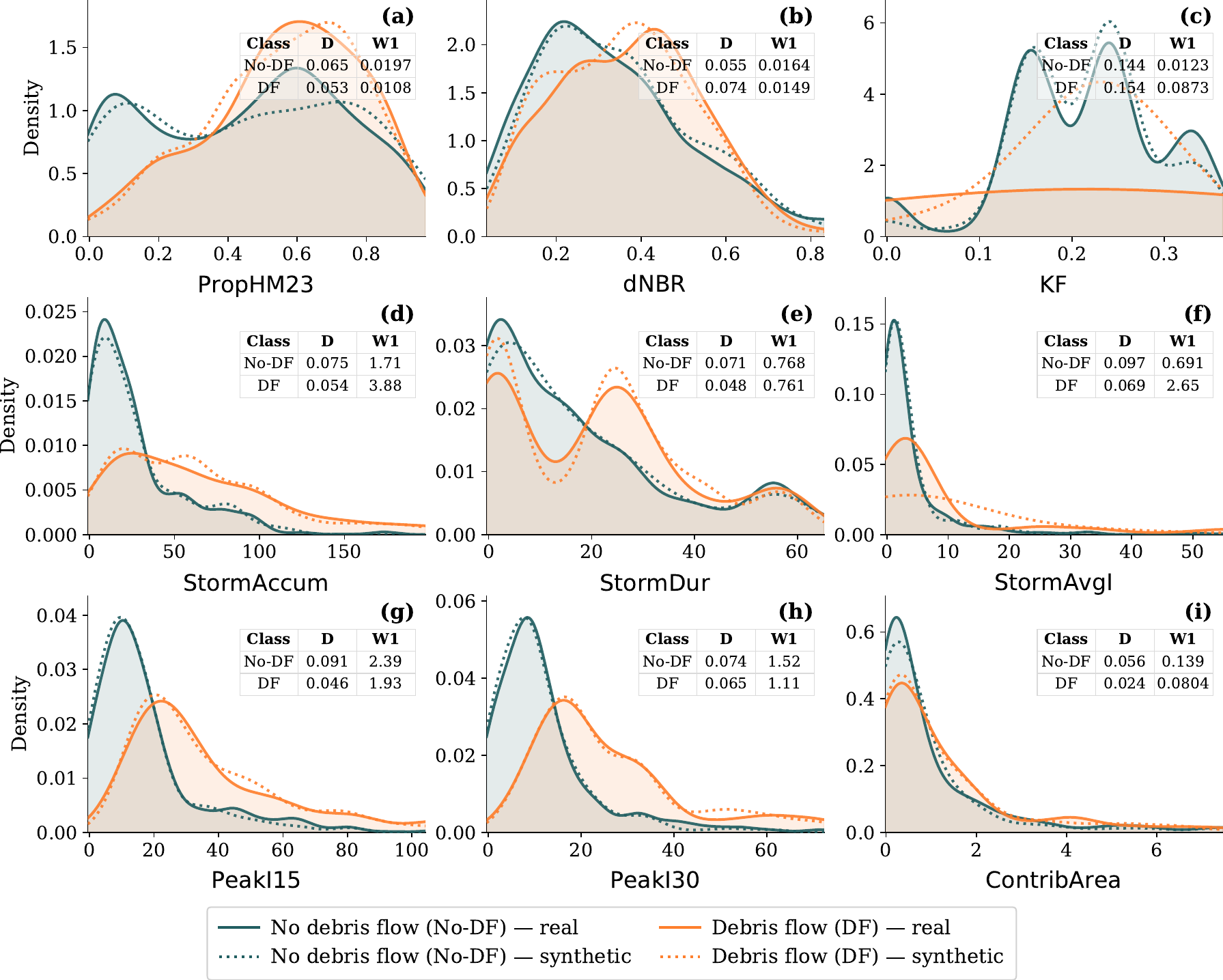}
    \caption{Class-conditional kernel density estimates of the benchmark features in a nine-panel grid, with panels~(a)--(i) shown separately for each feature and a shared bottom legend.
    Solid curves are imputed real training observations and dotted curves are fold~0 synthetic observations from the TabPFN unsupervised generator, with teal denoting the no-debris-flow (no DF) class and orange denoting the debris-flow (DF) class.
    Inset tables report the two-sample Kolmogorov--Smirnov statistic $D$ (bounded in $[0,1]$) and the Wasserstein-1 distance $W_1$ between real and synthetic observations for each class.}
    \label{fig:synthetic_distributions}
\end{figure}

\begin{figure}
    \centering
    \includegraphics[width=\linewidth]{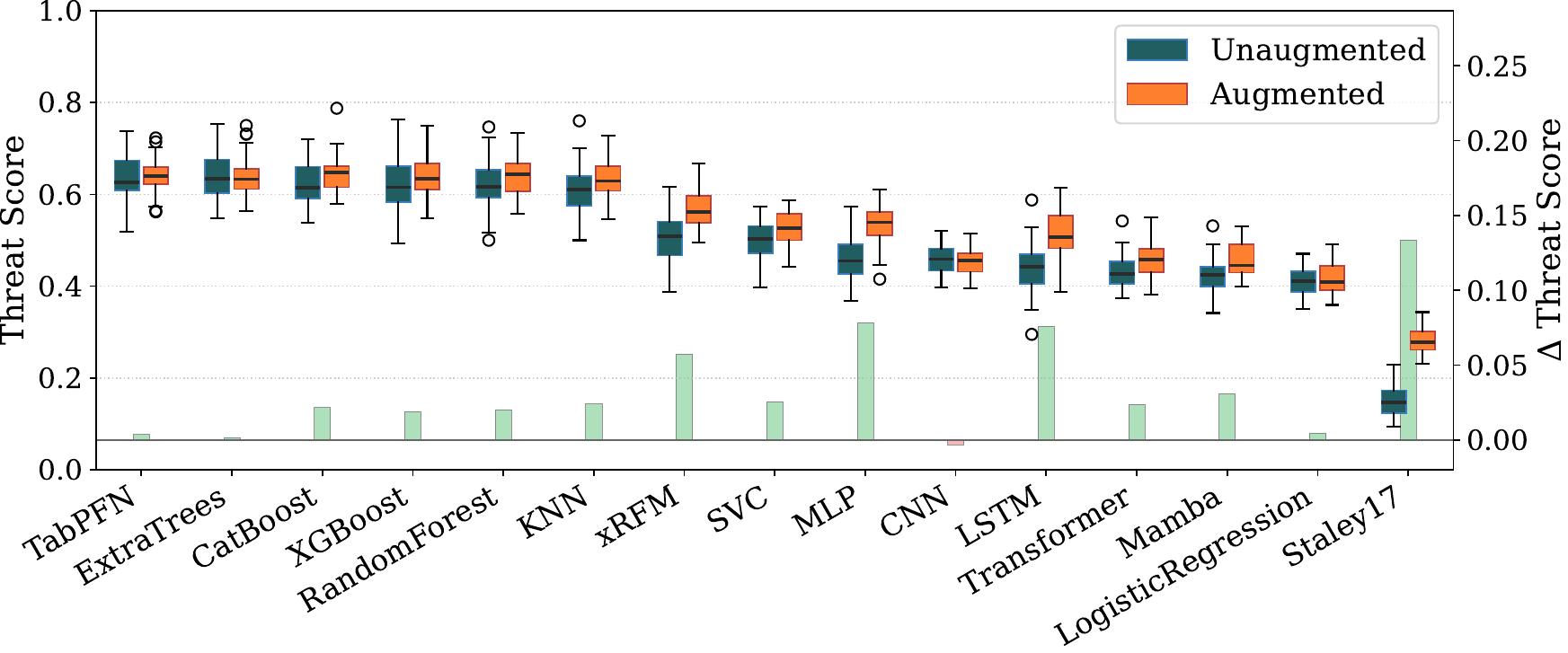}
    \caption{Per-model threat score for the unaugmented and augmented evaluations as overlaid box plots (left axis) and the mean change in threat score from unaugmented to augmented as bars (right axis).
    Models are ordered along the horizontal axis by descending unaugmented threat score to match the order in previous figures.}
    \label{fig:jaccard_delta}
\end{figure}

The synthetic data augmentation evaluation shows that TabPFN-generated synthetic rows can provide a measurable benefit for many models, but the effect is not uniform across models.
TabPFN serves two roles in this evaluation, as the generator of synthetic rows and as one of the evaluated models, which raises the possibility that augmentation could either favor TabPFN with synthetic rows that match its expected distribution or, conversely, leave it unchanged because TabPFN can already represent that distribution from the unaugmented training data alone.
The KDE and t-SNE realism checks (Figures~\ref{fig:synthetic_distributions} and~\ref{fig:synthetic_tsne}) confirm the synthetic rows are distributionally close to real, which narrows the dual-role question to whether TabPFN benefits more than other models from data it can already represent.

The observed pattern is that TabPFN does not benefit.
TabPFN gained only $+0.004$ threat score while several non-TabPFN models gained substantially more (Figure~\ref{fig:jaccard_delta}), which argues against a generator-model bias and is instead consistent with TabPFN already saturating the dataset's information.
That said, per-feature realism is uneven in a way that bears on interpretation: soil erodibility (\textit{KF}) carries the largest realism gap while rainfall features remain close, and \textit{KF} is also the feature whose SHAP ranking diverges most across models (rank 3--4 for XGBoost and CatBoost versus 10 for Random Forest), so any model that relies on a sharp \textit{KF} threshold may be the most exposed to residual generator error.
Such model-specific sensitivity to augmentation quality is expected in imbalanced classification, where resampling can improve minority sensitivity yet still interact with model bias and decision-boundary geometry \citep{chawla_smote_2002, he_learning_2009, branco_survey_2016}.

Gains in the augmented top tier (Table~\ref{tab:comparison_augmented}) are marginal and uneven.
CatBoost, Random Forest, and XGBoost each gained roughly two percentage points in threat score, while TabPFN and ExtraTrees moved by less than half a point, consistent with the unaugmented top tier already approaching the dataset's predictive ceiling.

Practically, synthetic data augmentation should be treated as a tunable component of the pipeline rather than a universal improvement step.
The strongest deployment strategy is to validate synthetic data augmentation choices on held-out real folds, report delta metrics, and retain only settings that improve both event detection and error trade-offs for the intended warning objective.

\section{Conclusion}
This study presented a unified evaluation of 15 models spanning five model families for post-wildfire debris-flow prediction across the western United States, to our knowledge, the first benchmark for this problem to include the new foundation model TabPFN, evaluate SHAP feature importance, and evaluate synthetic data augmentation.
TabPFN and the leading tree-based models (ExtraTrees, CatBoost, XGBoost, Random Forest) clustered tightly at threat score 0.62--0.64 against 0.15 for the Staley17 model, a roughly fourfold gap, with TabPFN matching the tuned models while requiring no dataset-specific hyperparameter search.
SHAP analyses across the four explained models converged on the same three highest-ranked and most fold-stable rainfall features (\textit{PeakI15}, \textit{StormAccum}, and \textit{PeakI30}), indicating that these model families emphasized the same predictive signals rather than divergent ones.
The four features Staley17 used to define its logistic regression model mapped only partially onto the SHAP rankings of the modern models.
Basin-level SHAP maps for two illustrative fires further showed spatial SHAP fields that tracked the corresponding feature fields and monotonic SHAP--feature dependence scatter relationships, indicating that TabPFN conditioned its predictions on feature--response behavior with the expected sign at those events rather than on spurious associations.
TabPFN-generated synthetic data augmentation, applied to mitigate the small, class-imbalanced training data, yielded the largest gain for the most restricted model, Staley17, 13.3 percentage points in threat score, while the top nonlinear tier improved by at most 2.2 percentage points, indicating that the benefit of synthetic augmentation was governed primarily by feature-set restriction and model capacity rather than acting as a universal enhancement.
These findings were most directly applicable to datasets with similar spatial and temporal scope, because model rankings were estimated under random stratified cross-validation and the data were concentrated in southern California within the first two post-wildfire years.
Repeated cross-validation strengthened ranking reliability relative to a single split, but it did not replace the need for stricter spatial transfer tests when models are deployed across new fire regions.
Future work should extend evaluation to multi-region datasets with strict spatial holdouts, longer post-wildfire temporal windows, and explicit threshold-optimization and cost-sensitive objectives that reflect the asymmetric consequences of missed events versus false alarms.

\FloatBarrier
\bibliographystyle{elsarticle-harv}
\bibliography{Cite}

@article{staley_prediction_2017,
	title = {Prediction of spatially explicit rainfall intensity–duration thresholds for post-fire debris-flow generation in the western United States},
	volume = {278},
	issn = {0169555X},
	url = {https://linkinghub.elsevier.com/retrieve/pii/S0169555X16302756},
	doi = {10.1016/j.geomorph.2016.10.019},
	pages = {149--162},
	journaltitle = {Geomorphology},
	shortjournal = {Geomorphology},
	author = {Staley, Dennis M. and Negri, Jacquelyn A. and Kean, Jason W. and Laber, Jayme L. and Tillery, Anne C. and Youberg, Ann M.},
	urldate = {2025-09-08},
	date = {2017-02},
	langid = {english},
}

@article{kern_machine_2017,
	title = {Machine Learning Based Predictive Modeling of Debris Flow Probability Following Wildfire in the Intermountain Western United States},
	volume = {49},
	issn = {1874-8961, 1874-8953},
	url = {http://link.springer.com/10.1007/s11004-017-9681-2},
	doi = {10.1007/s11004-017-9681-2},
	pages = {717--735},
	number = {6},
	journaltitle = {Mathematical Geosciences},
	shortjournal = {Math Geosci},
	author = {Kern, Ashley N. and Addison, Priscilla and Oommen, Thomas and Salazar, Sean E. and Coffman, Richard A.},
	urldate = {2025-09-03},
	date = {2017-08},
	langid = {english},
}

@report{staley_updated_2016,
	title = {Updated logistic regression equations for the calculation of post-fire debris-flow likelihood in the western United States},
	issn = {2331-1258},
	url = {https://pubs.usgs.gov/publication/ofr20161106},
	doi = {10.3133/ofr20161106},
	number = {2016-1106},
	institution = {U.S. Geological Survey},
	author = {Staley, Dennis M. and Negri, Jacquelyn A. and Kean, Jason W. and Laber, Jayme L. and Tillery, Anne C. and Youberg, Ann M.},
	urldate = {2025-09-24},
	date = {2016},
	langid = {english},
	note = {Publication Title: Open-File Report},
}

@article{nikolopoulos_evaluation_2018,
	title = {Evaluation of predictive models for post-fire debris flow occurrence in the western United States},
	volume = {18},
	rights = {https://creativecommons.org/licenses/by/4.0/},
	issn = {1684-9981},
	url = {https://nhess.copernicus.org/articles/18/2331/2018/},
	doi = {10.5194/nhess-18-2331-2018},
	pages = {2331--2343},
	number = {9},
	journaltitle = {Natural Hazards and Earth System Sciences},
	shortjournal = {Nat. Hazards Earth Syst. Sci.},
	author = {Nikolopoulos, Efthymios I. and Destro, Elisa and Bhuiyan, Md Abul Ehsan and Borga, Marco and Anagnostou, Emmanouil N.},
	urldate = {2025-09-24},
	date = {2018-09-04},
	langid = {english},
}

@inproceedings{roten_machine_2022,
	title = {Machine Learning for Improved Post-fire Debris Flow Likelihood Prediction},
	url = {https://ieeexplore.ieee.org/abstract/document/10020574},
	doi = {10.1109/BigData55660.2022.10020574},
	eventtitle = {2022 {IEEE} International Conference on Big Data (Big Data)},
	pages = {1681--1690},
	booktitle = {2022 {IEEE} International Conference on Big Data (Big Data)},
	author = {Roten, Daniel and Block, Jessica and Crawl, Daniel and Lee, Jenny and Altintas, Ilkay},
	urldate = {2025-09-24},
	date = {2022-12},
}

@article{cannon_predicting_2010,
	title = {Predicting the probability and volume of postwildfire debris flows in the intermountain western United States},
	volume = {122},
	issn = {0016-7606},
	url = {https://doi.org/10.1130/B26459.1},
	doi = {10.1130/B26459.1},
	pages = {127--144},
	number = {1},
	journaltitle = {{GSA} Bulletin},
	shortjournal = {{GSA} Bulletin},
	author = {Cannon, Susan H. and Gartner, Joseph E. and Rupert, Michael G. and Michael, John A. and Rea, Alan H. and Parrett, Charles},
	urldate = {2025-09-26},
	date = {2010-01-01},
}

@misc{grinsztajn_tabpfn-25_2025,
	title = {{TabPFN}-2.5: Advancing the State of the Art in Tabular Foundation Models},
	url = {http://arxiv.org/abs/2511.08667},
	doi = {10.48550/arXiv.2511.08667},
	shorttitle = {{TabPFN}-2.5},
	number = {{arXiv}:2511.08667},
	publisher = {{arXiv}},
	author = {Grinsztajn, Léo and Flöge, Klemens and Key, Oscar and Birkel, Felix and Jund, Philipp and Roof, Brendan and Jäger, Benjamin and Safaric, Dominik and Alessi, Simone and Hayler, Adrian and Manium, Mihir and Yu, Rosen and Jablonski, Felix and Hoo, Shi Bin and Garg, Anurag and Robertson, Jake and Bühler, Magnus and Moroshan, Vladyslav and Purucker, Lennart and Cornu, Clara and Wehrhahn, Lilly Charlotte and Bonetto, Alessandro and Schölkopf, Bernhard and Gambhir, Sauraj and Hollmann, Noah and Hutter, Frank},
	urldate = {2025-12-09},
	date = {2025-11-11},
	eprinttype = {arxiv},
	eprint = {2511.08667 [cs]},
}

@article{hollmann_accurate_2025,
	title = {Accurate predictions on small data with a tabular foundation model},
	volume = {637},
	rights = {2025 The Author(s)},
	issn = {1476-4687},
	url = {https://www.nature.com/articles/s41586-024-08328-6},
	doi = {10.1038/s41586-024-08328-6},
	pages = {319--326},
	number = {8045},
	journaltitle = {Nature},
	publisher = {Nature Publishing Group},
	author = {Hollmann, Noah and Müller, Samuel and Purucker, Lennart and Krishnakumar, Arjun and Körfer, Max and Hoo, Shi Bin and Schirrmeister, Robin Tibor and Hutter, Frank},
	urldate = {2025-12-17},
	date = {2025-01},
	langid = {english},
}

@misc{hollmann_tabpfn_2023,
	title = {{TabPFN}: A Transformer That Solves Small Tabular Classification Problems in a Second},
	url = {http://arxiv.org/abs/2207.01848},
	doi = {10.48550/arXiv.2207.01848},
	shorttitle = {{TabPFN}},
	number = {{arXiv}:2207.01848},
	publisher = {{arXiv}},
	author = {Hollmann, Noah and Müller, Samuel and Eggensperger, Katharina and Hutter, Frank},
	urldate = {2025-12-17},
	date = {2023-09-16},
	eprinttype = {arxiv},
	eprint = {2207.01848 [cs]},
}

@misc{beaglehole_xrfm_2025,
	title = {{xRFM}: Accurate, scalable, and interpretable feature learning models for tabular data},
	url = {http://arxiv.org/abs/2508.10053},
	doi = {10.48550/arXiv.2508.10053},
	shorttitle = {{xRFM}},
	number = {{arXiv}:2508.10053},
	publisher = {{arXiv}},
	author = {Beaglehole, Daniel and Holzmüller, David and Radhakrishnan, Adityanarayanan and Belkin, Mikhail},
	urldate = {2025-12-17},
	date = {2025-10-23},
	eprinttype = {arxiv},
	eprint = {2508.10053 [cs]},
}

@article{zhou_landslide_2025,
	title = {Landslide Risk Assessment as a Reference for Disaster Prevention and Mitigation: A Case Study of the Renhe District, Panzhihua City, China},
	volume = {17},
	issn = {2072-4292},
	url = {https://www.mdpi.com/2072-4292/17/13/2120},
	doi = {10.3390/rs17132120},
	shorttitle = {Landslide Risk Assessment as a Reference for Disaster Prevention and Mitigation},
	pages = {2120},
	number = {13},
	journaltitle = {Remote Sensing},
	shortjournal = {Remote Sensing},
	author = {Zhou, Yimeng and Xue, Lei and Ding, Hao and Wang, Haoyu and Huang, Kun and Li, Longfei and Li, Zhuan},
	urldate = {2026-01-14},
	date = {2025-06-20},
	langid = {english},
}

@article{addison_assessment_2019,
	title = {Assessment of post-wildfire debris flow occurrence using classifier tree},
	volume = {10},
	issn = {1947-5705},
	url = {https://doi.org/10.1080/19475705.2018.1530306},
	doi = {10.1080/19475705.2018.1530306},
	pages = {505--518},
	number = {1},
	journaltitle = {Geomatics, Natural Hazards and Risk},
	publisher = {Taylor \& Francis},
	author = {Addison, Priscilla and Oommen, Thomas and Sha, Qiuying},
	urldate = {2026-01-19},
	date = {2019-01-01},
	note = {\_eprint: https://doi.org/10.1080/19475705.2018.1530306},
}

@incollection{rundel_interpretable_2024,
	title = {Interpretable Machine Learning for {TabPFN}},
	volume = {2154},
	url = {http://arxiv.org/abs/2403.10923},
	doi = {10.1007/978-3-031-63797-1_23},
	pages = {465--476},
	author = {Rundel, David and Kobialka, Julius and Crailsheim, Constantin von and Feurer, Matthias and Nagler, Thomas and Rügamer, David},
	urldate = {2026-02-17},
	date = {2024},
	eprinttype = {arxiv},
	eprint = {2403.10923 [cs]},
}

@misc{muller_transformers_2024,
	title = {Transformers Can Do Bayesian Inference},
	url = {http://arxiv.org/abs/2112.10510},
	doi = {10.48550/arXiv.2112.10510},
	number = {{arXiv}:2112.10510},
	publisher = {{arXiv}},
	author = {Müller, Samuel and Hollmann, Noah and Arango, Sebastian Pineda and Grabocka, Josif and Hutter, Frank},
	urldate = {2026-02-20},
	date = {2024-08-13},
	eprinttype = {arxiv},
	eprint = {2112.10510 [cs]},
}

@article{staley_objective_2013,
	title = {Objective definition of rainfall intensity–duration thresholds for the initiation of post-fire debris flows in southern California},
	volume = {10},
	issn = {1612-5118},
	url = {https://doi.org/10.1007/s10346-012-0341-9},
	doi = {10.1007/s10346-012-0341-9},
	pages = {547--562},
	number = {5},
	journaltitle = {Landslides},
	shortjournal = {Landslides},
	author = {Staley, Dennis M. and Kean, Jason W. and Cannon, Susan H. and Schmidt, Kevin M. and Laber, Jayme L.},
	urldate = {2026-02-24},
	date = {2013-10-01},
	langid = {english},
}

@article{key_landscape_2006,
	title = {Landscape Assessment ({LA})},
	volume = {164},
	url = {https://research.fs.usda.gov/treesearch/24066},
	journaltitle = {In: Lutes, Duncan C.; Keane, Robert E.; Caratti, John F.; Key, Carl H.; Benson, Nathan C.; Sutherland, Steve; Gangi, Larry J. 2006. {FIREMON}: Fire effects monitoring and inventory system. Gen. Tech. Rep. {RMRS}-{GTR}-164-{CD}. Fort Collins, {CO}: U.S. Department of Agriculture, Forest Service, Rocky Mountain Research Station. p. {LA}-1-55},
	author = {Key, Carl H. and Benson, Nathan C.},
	urldate = {2026-02-24},
	date = {2006},
	langid = {english},
}

@misc{lundberg_unified_2017,
	title = {A Unified Approach to Interpreting Model Predictions},
	url = {http://arxiv.org/abs/1705.07874},
	doi = {10.48550/arXiv.1705.07874},
	number = {{arXiv}:1705.07874},
	publisher = {{arXiv}},
	author = {Lundberg, Scott and Lee, Su-In},
	urldate = {2026-02-27},
	date = {2017-11-25},
	eprinttype = {arxiv},
	eprint = {1705.07874 [cs]},
}

@article{zhu_method_2026,
	title = {A method for better mapping of susceptibility to thaw hazards in data-scarce cold regions},
	volume = {337},
	issn = {0034-4257},
	url = {https://www.sciencedirect.com/science/article/pii/S0034425726001082},
	doi = {10.1016/j.rse.2026.115338},
	pages = {115338},
	journaltitle = {Remote Sensing of Environment},
	shortjournal = {Remote Sensing of Environment},
	author = {Zhu, Hualiang and Zhang, Xianwei and Wei, Gang and Wang, Qingzhi and Liu, Xinyu and Yan, Lei and Wang, Gang},
	urldate = {2026-03-10},
	date = {2026-05-01},
}

@article{sattele_reliability_2015,
	title = {Reliability and effectiveness of early warning systems for natural hazards: Concept and application to debris flow warning},
	volume = {142},
	issn = {0951-8320},
	url = {https://www.sciencedirect.com/science/article/pii/S0951832015001441},
	doi = {10.1016/j.ress.2015.05.003},
	shorttitle = {Reliability and effectiveness of early warning systems for natural hazards},
	pages = {192--202},
	journaltitle = {Reliability Engineering \& System Safety},
	shortjournal = {Reliability Engineering \& System Safety},
	author = {Sättele, Martina and Bründl, Michael and Straub, Daniel},
	urldate = {2026-04-30},
	date = {2015-10-01},
}

@online{yang_knowledge-data_2026,
	title = {Knowledge-Data Dually Driven Paradigm for Accurate Landslide Susceptibility Prediction under Data-Scarce Conditions Using Geomorphic Priors and Tabular Foundation Model},
	url = {https://arxiv.org/abs/2604.25196v1},
	titleaddon = {{arXiv}.org},
	author = {Yang, Yuting and Mei, Gang and Chen, Feng and Zhang, Yongshuang and Peng, Jianbing},
	urldate = {2026-05-07},
	date = {2026-04-28},
	langid = {english},
}

@article{jin_susceptibility_2022,
	title = {Susceptibility Prediction of Post-Fire Debris Flows in Xichang, China, Using a Logistic Regression Model from a Spatiotemporal Perspective},
	volume = {14},
	rights = {http://creativecommons.org/licenses/by/3.0/},
	issn = {2072-4292},
	url = {https://www.mdpi.com/2072-4292/14/6/1306},
	doi = {10.3390/rs14061306},
	pages = {1306},
	number = {6},
	journaltitle = {Remote Sensing},
	publisher = {Multidisciplinary Digital Publishing Institute},
	author = {Jin, Tao and Hu, Xiewen and Liu, Bo and Xi, Chuanjie and He, Kun and Cao, Xichao and Luo, Gang and Han, Mei and Ma, Guotao and Yang, Ying and Wang, Yan},
	urldate = {2026-05-08},
	date = {2022-01},
	langid = {english},
}

@article{diakakis_exploring_2023,
	title = {Exploring the Application of a Debris Flow Likelihood Regression Model in Mediterranean Post-Fire Environments, Using Field Observations-Based Validation},
	volume = {12},
	rights = {http://creativecommons.org/licenses/by/3.0/},
	issn = {2073-445X},
	url = {https://www.mdpi.com/2073-445X/12/3/555},
	doi = {10.3390/land12030555},
	pages = {555},
	number = {3},
	journaltitle = {Land},
	publisher = {Multidisciplinary Digital Publishing Institute},
	author = {Diakakis, Michalis and Mavroulis, Spyridon and Vassilakis, Emmanuel and Chalvatzi, Vassiliki},
	urldate = {2026-05-08},
	date = {2023-03},
	langid = {english},
}

@article{kumar_addressing_2024,
	title = {Addressing class imbalance in soil movement predictions},
	volume = {24},
	issn = {1561-8633},
	url = {https://nhess.copernicus.org/articles/24/1913/2024/},
	doi = {10.5194/nhess-24-1913-2024},
	pages = {1913--1928},
	number = {6},
	journaltitle = {Natural Hazards and Earth System Sciences},
	publisher = {Copernicus {GmbH}},
	author = {Kumar, Praveen and Priyanka, Priyanka and Uday, Kala Venkata and Dutt, Varun},
	urldate = {2026-05-11},
	date = {2024-06-06},
}

@article{wang_optimizing_2019,
	title = {Optimizing the Predictive Ability of Machine Learning Methods for Landslide Susceptibility Mapping Using {SMOTE} for Lishui City in Zhejiang Province, China},
	volume = {16},
	rights = {http://creativecommons.org/licenses/by/3.0/},
	issn = {1660-4601},
	url = {https://www.mdpi.com/1660-4601/16/3/368},
	doi = {10.3390/ijerph16030368},
	pages = {368},
	number = {3},
	journaltitle = {International Journal of Environmental Research and Public Health},
	publisher = {Multidisciplinary Digital Publishing Institute},
	author = {Wang, Yumiao and Wu, Xueling and Chen, Zhangjian and Ren, Fu and Feng, Luwei and Du, Qingyun},
	urldate = {2026-05-11},
	date = {2019-01},
	langid = {english},
}

@article{hanley_meaning_1982,
	title = {The meaning and use of the area under a receiver operating characteristic ({ROC}) curve.},
	volume = {143},
	url = {https://doi.org/10.1148/radiology.143.1.7063747},
	doi = {10.1148/radiology.143.1.7063747},
	pages = {29--36},
	number = {1},
	journaltitle = {Radiology},
	author = {Hanley, J A and {McNeil}, B J},
	date = {1982},
	note = {\_eprint: https://doi.org/10.1148/radiology.143.1.7063747},
}

@inproceedings{chen_xgboost_2016,
	location = {New York, {NY}, {USA}},
	title = {{XGBoost}: A Scalable Tree Boosting System},
	isbn = {978-1-4503-4232-2},
	url = {https://dl.acm.org/doi/10.1145/2939672.2939785},
	doi = {10.1145/2939672.2939785},
	series = {{KDD} '16},
	shorttitle = {{XGBoost}},
	pages = {785--794},
	booktitle = {Proceedings of the 22nd {ACM} {SIGKDD} International Conference on Knowledge Discovery and Data Mining},
	publisher = {Association for Computing Machinery},
	author = {Chen, Tianqi and Guestrin, Carlos},
	urldate = {2026-05-11},
	date = {2016-08-13},
}

@article{orland_scalable_2022,
	title = {A Scalable Framework for Post Fire Debris Flow Hazard Assessment Using Satellite Precipitation Data},
	volume = {49},
	rights = {© 2022. The Authors.},
	issn = {1944-8007},
	url = {https://onlinelibrary.wiley.com/doi/abs/10.1029/2022GL099850},
	doi = {10.1029/2022GL099850},
	pages = {e2022GL099850},
	number = {18},
	journaltitle = {Geophysical Research Letters},
	author = {Orland, Elijah and Kirschbaum, Dalia and Stanley, Thomas},
	urldate = {2026-05-11},
	date = {2022},
	langid = {english},
	note = {\_eprint: https://agupubs.onlinelibrary.wiley.com/doi/pdf/10.1029/2022GL099850},
}

@inproceedings{louppe_understanding_2013,
	title = {Understanding variable importances in forests of randomized trees},
	volume = {26},
	url = {https://proceedings.neurips.cc/paper_files/paper/2013/hash/e3796ae838835da0b6f6ea37bcf8bcb7-Abstract.html},
	booktitle = {Advances in Neural Information Processing Systems},
	publisher = {Curran Associates, Inc.},
	author = {Louppe, Gilles and Wehenkel, Louis and Sutera, Antonio and Geurts, Pierre},
	urldate = {2026-05-11},
	date = {2013},
}

@article{breiman_random_2001,
	title = {Random Forests},
	volume = {45},
	issn = {1573-0565},
	url = {https://doi.org/10.1023/A:1010933404324},
	doi = {10.1023/A:1010933404324},
	pages = {5--32},
	number = {1},
	journaltitle = {Machine Learning},
	shortjournal = {Machine Learning},
	author = {Breiman, Leo},
	urldate = {2026-05-11},
	date = {2001-10-01},
	langid = {english},
}

@article{xu_comparative_2014,
	title = {A comparative study of different classification techniques for marine oil spill identification using {RADARSAT}-1 imagery},
	volume = {141},
	issn = {0034-4257},
	url = {https://www.sciencedirect.com/science/article/pii/S0034425713003805},
	doi = {10.1016/j.rse.2013.10.012},
	pages = {14--23},
	journaltitle = {Remote Sensing of Environment},
	shortjournal = {Remote Sensing of Environment},
	author = {Xu, Linlin and Li, Jonathan and Brenning, Alexander},
	urldate = {2026-05-11},
	date = {2014-02-05},
}

@article{chawla_smote_2002,
	title = {{SMOTE}: Synthetic Minority Over-sampling Technique},
	volume = {16},
	rights = {Copyright (c)},
	issn = {1076-9757},
	url = {https://www.jair.org/index.php/jair/article/view/10302},
	doi = {10.1613/jair.953},
	shorttitle = {{SMOTE}},
	pages = {321--357},
	journaltitle = {Journal of Artificial Intelligence Research},
	author = {Chawla, N. V. and Bowyer, K. W. and Hall, L. O. and Kegelmeyer, W. P.},
	urldate = {2026-05-11},
	date = {2002-06-01},
	langid = {english},
}

@article{he_learning_2009,
	title = {Learning from Imbalanced Data},
	volume = {21},
	issn = {1558-2191},
	url = {https://ieeexplore.ieee.org/document/5128907},
	doi = {10.1109/TKDE.2008.239},
	pages = {1263--1284},
	number = {9},
	journaltitle = {{IEEE} Transactions on Knowledge and Data Engineering},
	author = {He, Haibo and Garcia, Edwardo A.},
	urldate = {2026-05-11},
	date = {2009-09},
}

@article{branco_survey_2016,
	title = {A Survey of Predictive Modeling on Imbalanced Domains},
	volume = {49},
	issn = {0360-0300},
	url = {https://dl.acm.org/doi/10.1145/2907070},
	doi = {10.1145/2907070},
	pages = {31:1--31:50},
	number = {2},
	journaltitle = {{ACM} Comput. Surv.},
	author = {Branco, Paula and Torgo, Luís and Ribeiro, Rita P.},
	urldate = {2026-05-11},
	date = {2016-08-13},
}

@article{lecun_gradient-based_1998,
	title = {Gradient-based learning applied to document recognition},
	volume = {86},
	issn = {1558-2256},
	url = {https://ieeexplore.ieee.org/document/726791},
	doi = {10.1109/5.726791},
	pages = {2278--2324},
	number = {11},
	journaltitle = {Proceedings of the {IEEE}},
	author = {Lecun, Y. and Bottou, L. and Bengio, Y. and Haffner, P.},
	urldate = {2026-05-11},
	date = {1998-11},
}

@inproceedings{akiba_optuna_2019,
	location = {New York, {NY}, {USA}},
	title = {Optuna: A Next-generation Hyperparameter Optimization Framework},
	isbn = {978-1-4503-6201-6},
	url = {https://dl.acm.org/doi/10.1145/3292500.3330701},
	doi = {10.1145/3292500.3330701},
	series = {{KDD} '19},
	shorttitle = {Optuna},
	pages = {2623--2631},
	booktitle = {Proceedings of the 25th {ACM} {SIGKDD} International Conference on Knowledge Discovery \& Data Mining},
	publisher = {Association for Computing Machinery},
	author = {Akiba, Takuya and Sano, Shotaro and Yanase, Toshihiko and Ohta, Takeru and Koyama, Masanori},
	urldate = {2026-05-11},
	date = {2019-07-25},
}

@article{li_hyperband_2018,
	title = {Hyperband: A Novel Bandit-Based Approach to Hyperparameter Optimization},
	volume = {18},
	issn = {1533-7928},
	url = {http://jmlr.org/papers/v18/16-558.html},
	shorttitle = {Hyperband},
	pages = {1--52},
	number = {185},
	journaltitle = {Journal of Machine Learning Research},
	author = {Li, Lisha and Jamieson, Kevin and {DeSalvo}, Giulia and Rostamizadeh, Afshin and Talwalkar, Ameet},
	urldate = {2026-05-11},
	date = {2018},
}

@article{maaten_visualizing_2008,
	title = {Visualizing Data using t-{SNE}},
	volume = {9},
	issn = {1533-7928},
	url = {http://jmlr.org/papers/v9/vandermaaten08a.html},
	pages = {2579--2605},
	number = {86},
	journaltitle = {Journal of Machine Learning Research},
	author = {Maaten, Laurens van der and Hinton, Geoffrey},
	urldate = {2026-05-11},
	date = {2008},
}

@article{cox_regression_1958,
	title = {The Regression Analysis of Binary Sequences},
	volume = {20},
	issn = {0035-9246},
	url = {https://doi.org/10.1111/j.2517-6161.1958.tb00292.x},
	doi = {10.1111/j.2517-6161.1958.tb00292.x},
	pages = {215--232},
	number = {2},
	journaltitle = {Journal of the Royal Statistical Society: Series B (Methodological)},
	shortjournal = {Royal Statistical Society. Journal. Series B: Methodological},
	author = {Cox, D. R.},
	urldate = {2026-05-11},
	date = {1958-07-01},
}

@incollection{villani_wasserstein_2009,
	location = {Berlin, Heidelberg},
	title = {The Wasserstein distances},
	isbn = {978-3-540-71050-9},
	url = {https://doi.org/10.1007/978-3-540-71050-9_6},
	doi = {10.1007/978-3-540-71050-9_6},
	pages = {93--111},
	booktitle = {Optimal Transport: Old and New},
	publisher = {Springer},
	author = {Villani, Cédric},
	editor = {Villani, Cédric},
	urldate = {2026-05-11},
	date = {2009},
	langid = {english},
}

@article{cover_nearest_1967,
	title = {Nearest neighbor pattern classification},
	volume = {13},
	issn = {1557-9654},
	url = {https://ieeexplore.ieee.org/document/1053964},
	doi = {10.1109/TIT.1967.1053964},
	pages = {21--27},
	number = {1},
	journaltitle = {{IEEE} Transactions on Information Theory},
	author = {Cover, T. and Hart, P.},
	urldate = {2026-05-11},
	date = {1967-01},
}

@article{cortes_support-vector_1995,
	title = {Support-vector networks},
	volume = {20},
	issn = {1573-0565},
	url = {https://doi.org/10.1007/BF00994018},
	doi = {10.1007/BF00994018},
	pages = {273--297},
	number = {3},
	journaltitle = {Machine Learning},
	shortjournal = {Mach Learn},
	author = {Cortes, Corinna and Vapnik, Vladimir},
	urldate = {2026-05-11},
	date = {1995-09-01},
	langid = {english},
}

@article{geurts_extremely_2006,
	title = {Extremely randomized trees},
	volume = {63},
	issn = {1573-0565},
	url = {https://doi.org/10.1007/s10994-006-6226-1},
	doi = {10.1007/s10994-006-6226-1},
	pages = {3--42},
	number = {1},
	journaltitle = {Machine Learning},
	shortjournal = {Mach Learn},
	author = {Geurts, Pierre and Ernst, Damien and Wehenkel, Louis},
	urldate = {2026-05-11},
	date = {2006-04-01},
	langid = {english},
}

@article{rumelhart_learning_1986,
	title = {Learning representations by back-propagating errors},
	volume = {323},
	rights = {1986 Springer Nature Limited},
	issn = {1476-4687},
	url = {https://www.nature.com/articles/323533a0},
	doi = {10.1038/323533a0},
	pages = {533--536},
	number = {6088},
	journaltitle = {Nature},
	publisher = {Nature Publishing Group},
	author = {Rumelhart, David E. and Hinton, Geoffrey E. and Williams, Ronald J.},
	urldate = {2026-05-11},
	date = {1986-10},
	langid = {english},
}

@article{szekely_energy_2013,
	title = {Energy statistics: A class of statistics based on distances},
	volume = {143},
	issn = {0378-3758},
	url = {https://www.sciencedirect.com/science/article/pii/S0378375813000633},
	doi = {10.1016/j.jspi.2013.03.018},
	shorttitle = {Energy statistics},
	pages = {1249--1272},
	number = {8},
	journaltitle = {Journal of Statistical Planning and Inference},
	shortjournal = {Journal of Statistical Planning and Inference},
	author = {Székely, Gábor J. and Rizzo, Maria L.},
	urldate = {2026-05-11},
	date = {2013-08-01},
}

@misc{grinsztajn_why_2022,
	title = {Why do tree-based models still outperform deep learning on tabular data?},
	url = {http://arxiv.org/abs/2207.08815},
	doi = {10.48550/arXiv.2207.08815},
	number = {{arXiv}:2207.08815},
	publisher = {{arXiv}},
	author = {Grinsztajn, Léo and Oyallon, Edouard and Varoquaux, Gaël},
	urldate = {2026-05-11},
	date = {2022-07-18},
	eprinttype = {arxiv},
	eprint = {2207.08815 [cs.LG]},
}

@article{borisov_deep_2024,
	title = {Deep Neural Networks and Tabular Data: A Survey},
	volume = {35},
	issn = {2162-2388},
	url = {https://ieeexplore.ieee.org/document/9998482},
	doi = {10.1109/TNNLS.2022.3229161},
	shorttitle = {Deep Neural Networks and Tabular Data},
	pages = {7499--7519},
	number = {6},
	journaltitle = {{IEEE} Transactions on Neural Networks and Learning Systems},
	author = {Borisov, Vadim and Leemann, Tobias and Seßler, Kathrin and Haug, Johannes and Pawelczyk, Martin and Kasneci, Gjergji},
	urldate = {2026-05-11},
	date = {2024-06},
}

@misc{xu_modeling_2019,
	title = {Modeling Tabular data using Conditional {GAN}},
	url = {http://arxiv.org/abs/1907.00503},
	doi = {10.48550/arXiv.1907.00503},
	number = {{arXiv}:1907.00503},
	publisher = {{arXiv}},
	author = {Xu, Lei and Skoularidou, Maria and Cuesta-Infante, Alfredo and Veeramachaneni, Kalyan},
	urldate = {2026-05-11},
	date = {2019-10-28},
	eprinttype = {arxiv},
	eprint = {1907.00503 [cs.LG]},
}

@article{yang_machine-learning-based_2024,
	title = {Machine-Learning-Based Prediction Modeling for Debris Flow Occurrence: A Meta-Analysis},
	volume = {16},
	rights = {http://creativecommons.org/licenses/by/3.0/},
	issn = {2073-4441},
	url = {https://www.mdpi.com/2073-4441/16/7/923},
	doi = {10.3390/w16070923},
	shorttitle = {Machine-Learning-Based Prediction Modeling for Debris Flow Occurrence},
	pages = {923},
	number = {7},
	journaltitle = {Water},
	publisher = {Multidisciplinary Digital Publishing Institute},
	author = {Yang, Lianbing and Ge, Yonggang and Chen, Baili and Wu, Yuhong and Fu, Runde},
	urldate = {2026-05-11},
	date = {2024-01},
	langid = {english},
}

@article{degraff_timing_2015,
	title = {The Timing of Susceptibility to Post-Fire Debris Flows in the Western United States},
	volume = {21},
	issn = {1078-7275},
	url = {https://doi.org/10.2113/gseegeosci.21.4.277},
	doi = {10.2113/gseegeosci.21.4.277},
	pages = {277--292},
	number = {4},
	journaltitle = {Environmental \& Engineering Geoscience},
	shortjournal = {Environmental \& Engineering Geoscience},
	author = {{DeGraff}, Jerome V. and Cannon, Susan H. and Gartner, Joseph E.},
	urldate = {2026-05-11},
	date = {2015-11-01},
}

@inproceedings{brown_language_2020,
	title = {Language Models are Few-Shot Learners},
	volume = {33},
	url = {https://papers.nips.cc/paper/2020/hash/1457c0d6bfcb4967418bfb8ac142f64a-Abstract.html},
	pages = {1877--1901},
	booktitle = {Advances in Neural Information Processing Systems},
	publisher = {Curran Associates, Inc.},
	author = {Brown, Tom and Mann, Benjamin and Ryder, Nick and Subbiah, Melanie and Kaplan, Jared D and Dhariwal, Prafulla and Neelakantan, Arvind and Shyam, Pranav and Sastry, Girish and Askell, Amanda and Agarwal, Sandhini and Herbert-Voss, Ariel and Krueger, Gretchen and Henighan, Tom and Child, Rewon and Ramesh, Aditya and Ziegler, Daniel and Wu, Jeffrey and Winter, Clemens and Hesse, Chris and Chen, Mark and Sigler, Eric and Litwin, Mateusz and Gray, Scott and Chess, Benjamin and Clark, Jack and Berner, Christopher and {McCandlish}, Sam and Radford, Alec and Sutskever, Ilya and Amodei, Dario},
	urldate = {2026-05-11},
	date = {2020},
}

@book{hastie_elements_2009,
	location = {New York, {NY}},
	title = {The Elements of Statistical Learning},
	rights = {http://www.springer.com/tdm},
	isbn = {978-0-387-84857-0 978-0-387-84858-7},
	url = {http://link.springer.com/10.1007/978-0-387-84858-7},
	doi = {10.1007/978-0-387-84858-7},
	series = {Springer Series in Statistics},
	publisher = {Springer New York},
	author = {Hastie, Trevor and Tibshirani, Robert and Friedman, Jerome},
	urldate = {2026-05-11},
	date = {2009},
}

@misc{gu_mamba_2024,
	title = {Mamba: Linear-Time Sequence Modeling with Selective State Spaces},
	url = {http://arxiv.org/abs/2312.00752},
	doi = {10.48550/arXiv.2312.00752},
	shorttitle = {Mamba},
	number = {{arXiv}:2312.00752},
	publisher = {{arXiv}},
	author = {Gu, Albert and Dao, Tri},
	urldate = {2026-05-11},
	date = {2024-05-31},
	eprinttype = {arxiv},
	eprint = {2312.00752 [cs.LG]},
}

@article{arlot_survey_2010,
	title = {A survey of cross-validation procedures for model selection},
	volume = {4},
	issn = {1935-7516},
	url = {https://projecteuclid.org/journals/statistics-surveys/volume-4/issue-none/A-survey-of-cross-validation-procedures-for-model-selection/10.1214/09-SS054.full},
	doi = {10.1214/09-SS054},
	issue = {none},
	journaltitle = {Statistics Surveys},
	shortjournal = {Statist. Surv.},
	author = {Arlot, Sylvain and Celisse, Alain},
	urldate = {2026-05-11},
	date = {2010-01-01},
}

@inproceedings{kohavi_study_1995,
	location = {San Francisco, {CA}, {USA}},
	title = {A study of cross-validation and bootstrap for accuracy estimation and model selection},
	isbn = {978-1-55860-363-9},
	series = {{IJCAI}'95},
	pages = {1137--1143},
	booktitle = {Proceedings of the 14th international joint conference on Artificial intelligence - Volume 2},
	publisher = {Morgan Kaufmann Publishers Inc.},
	author = {Kohavi, Ron},
	urldate = {2026-05-11},
	date = {1995-08-20},
}

@article{schwarz_state_1995,
	title = {State Soil Geographic ({STATSGO}) Data Base for the Conterminous United States},
	issn = {2331-1258},
	url = {https://pubs.usgs.gov/publication/ofr95449},
	doi = {10.3133/ofr95449},
	journaltitle = {Open-File Report},
	author = {Schwarz, Gregory E. and Alexander, R. B.},
	urldate = {2026-05-15},
	date = {1995},
	langid = {english},
	note = {Number: 95-449},
}

@article{cawley_over-fitting_2010,
	title = {On Over-fitting in Model Selection and Subsequent Selection Bias in Performance Evaluation},
	volume = {11},
	issn = {1532-4435},
	url = {https://dl.acm.org/doi/10.5555/1756006.1859921},
	pages = {2079--2107},
	journaltitle = {The Journal of Machine Learning Research},
	shortjournal = {J. Mach. Learn. Res.},
	author = {Cawley, Gavin C. and Talbot, Nicola L.C.},
	urldate = {2026-05-15},
	date = {2010-08-01},
}

@article{koutsoyiannis_mathematical_1998,
	title = {A mathematical framework for studying rainfall intensity-duration-frequency relationships},
	volume = {206},
	issn = {0022-1694},
	url = {https://www.sciencedirect.com/science/article/pii/S0022169498000973},
	doi = {10.1016/S0022-1694(98)00097-3},
	pages = {118--135},
	number = {1},
	journaltitle = {Journal of Hydrology},
	shortjournal = {Journal of Hydrology},
	author = {Koutsoyiannis, Demetris and Kozonis, Demosthenes and Manetas, Alexandros},
	urldate = {2026-05-18},
	date = {1998-04-01},
}

@software{grinsztajn_priorlabstabpfn-extensions_2026,
	title = {{PriorLabs}/tabpfn-extensions},
	url = {https://github.com/PriorLabs/tabpfn-extensions},
	version = {0.2.2},
	publisher = {Prior Labs},
	author = {Grinsztajn, Leo and Purucker, Lennart and Krishnakumar, Arjun and Körfer, Max and Hoo, Shi Bin and Schirrmeister, Robin Tibor and Bergman, Eddie and Flöge, Klemens and Key, Oscar and Safaric, Dominik and Hollmann, Noah and Müller, Samuel and Hutter, Frank},
	urldate = {2026-05-11},
	date = {2026-05-11},
}

@article{schaefer_critical_1990,
	title = {The Critical Success Index as an Indicator of Warning Skill},
	volume = {5},
	issn = {1520-0434, 0882-8156},
	url = {https://journals.ametsoc.org/view/journals/wefo/5/4/1520-0434_1990_005_0570_tcsiaa_2_0_co_2.xml},
	doi = {10.1175/1520-0434(1990)005<0570:TCSIAA>2.0.CO;2},
	pages = {570--575},
	number = {4},
	journaltitle = {Weather and Forecasting},
	publisher = {American Meteorological Society},
	author = {Schaefer, Joseph T.},
	urldate = {2026-05-26},
	date = {1990-12-01},
}

@software{jonathan_m_king_pfdf_2025,
	title = {pfdf - Python library for postfire debris-flow hazard assessments and research, version 3.0.2},
	url = {https://code.usgs.gov/ghsc/lhp/pfdf/-/releases/3.0.0},
	doi = {10.5066/P1JJXSXD},
	publisher = {U.S. Geological Survey},
	author = {Jonathan M King},
	urldate = {2026-05-26},
	date = {2025-03-07},
}

@misc{ribeiro_why_2016,
	title = {"Why Should I Trust You?": Explaining the Predictions of Any Classifier},
	url = {http://arxiv.org/abs/1602.04938},
	doi = {10.48550/arXiv.1602.04938},
	shorttitle = {"Why Should I Trust You?},
	number = {{arXiv}:1602.04938},
	publisher = {{arXiv}},
	author = {Ribeiro, Marco Tulio and Singh, Sameer and Guestrin, Carlos},
	urldate = {2026-05-26},
	date = {2016-08-09},
	eprinttype = {arxiv},
	eprint = {1602.04938 [cs.LG]},
}

@misc{lundberg_explainable_2019,
	title = {Explainable {AI} for Trees: From Local Explanations to Global Understanding},
	url = {http://arxiv.org/abs/1905.04610},
	doi = {10.48550/arXiv.1905.04610},
	shorttitle = {Explainable {AI} for Trees},
	number = {{arXiv}:1905.04610},
	publisher = {{arXiv}},
	author = {Lundberg, Scott M. and Erion, Gabriel and Chen, Hugh and {DeGrave}, Alex and Prutkin, Jordan M. and Nair, Bala and Katz, Ronit and Himmelfarb, Jonathan and Bansal, Nisha and Lee, Su-In},
	urldate = {2026-05-26},
	date = {2019-05-11},
	eprinttype = {arxiv},
	eprint = {1905.04610 [cs.LG]},
}

@article{demsar_statistical_2006,
	title = {Statistical Comparisons of Classifiers over Multiple Data Sets},
	volume = {7},
	issn = {1533-7928},
	url = {http://jmlr.org/papers/v7/demsar06a.html},
	pages = {1--30},
	number = {1},
	journaltitle = {Journal of Machine Learning Research},
	author = {Demšar, Janez},
	urldate = {2026-05-27},
	date = {2006},
}

@article{hochreiter_long_1997,
	title = {Long short-term memory.},
	volume = {9},
	issn = {0899-7667},
	url = {https://research.ebsco.com/plink/f5d31564-c30a-3dc4-b6a8-685d81bfe506},
	doi = {10.1162/neco.1997.9.8.1735},
	pages = {1735},
	number = {8},
	journaltitle = {Neural Computation},
	shortjournal = {Neural Computation},
	publisher = {{MIT} Press},
	author = {Hochreiter, Sepp and Schmidhuber, Jurgen},
	urldate = {2026-05-27},
	date = {1997-11-15},
}

@article{massey_kolmogorov-smirnov_1951,
	title = {The Kolmogorov-Smirnov Test for Goodness of Fit},
	volume = {46},
	issn = {0162-1459},
	url = {https://www.jstor.org/stable/2280095},
	doi = {10.2307/2280095},
	pages = {68--78},
	number = {253},
	journaltitle = {Journal of the American Statistical Association},
	publisher = {[American Statistical Association, Taylor \& Francis, Ltd.]},
	author = {Massey, Frank J.},
	urldate = {2026-05-27},
	date = {1951},
}

@article{krstajic_cross-validation_2014,
	title = {Cross-validation pitfalls when selecting and assessing regression and classification models},
	volume = {6},
	issn = {1758-2946},
	url = {https://pmc.ncbi.nlm.nih.gov/articles/PMC3994246/},
	doi = {10.1186/1758-2946-6-10},
	pages = {10},
	journaltitle = {Journal of Cheminformatics},
	shortjournal = {J Cheminform},
	author = {Krstajic, Damjan and Buturovic, Ljubomir J and Leahy, David E and Thomas, Simon},
	urldate = {2026-05-27},
	date = {2014-03-29},
}

@inproceedings{vaswani_attention_2017,
	title = {Attention is All you Need},
	volume = {30},
	url = {https://proceedings.neurips.cc/paper_files/paper/2017/hash/3f5ee243547dee91fbd053c1c4a845aa-Abstract.html},
	booktitle = {Advances in Neural Information Processing Systems},
	publisher = {Curran Associates, Inc.},
	author = {Vaswani, Ashish and Shazeer, Noam and Parmar, Niki and Uszkoreit, Jakob and Jones, Llion and Gomez, Aidan N and Kaiser, {\L}ukasz and Polosukhin, Illia},
	urldate = {2026-05-27},
	date = {2017},
}

@inproceedings{prokhorenkova_catboost_2018,
	title = {{CatBoost}: unbiased boosting with categorical features},
	volume = {31},
	url = {https://proceedings.neurips.cc/paper_files/paper/2018/hash/14491b756b3a51daac41c24863285549-Abstract.html},
	shorttitle = {{CatBoost}},
	booktitle = {Advances in Neural Information Processing Systems},
	publisher = {Curran Associates, Inc.},
	author = {Prokhorenkova, Liudmila and Gusev, Gleb and Vorobev, Aleksandr and Dorogush, Anna Veronika and Gulin, Andrey},
	urldate = {2026-05-27},
	date = {2018},
}

@article{dietterich_approximate_1998,
	title = {Approximate Statistical Tests for Comparing Supervised Classification Learning Algorithms.},
	volume = {10},
	issn = {0899-7667},
	url = {https://research.ebsco.com/plink/e22f08e8-ee48-37c6-975c-436955c10e29},
	doi = {10.1162/089976698300017197},
	pages = {1895--1923},
	number = {7},
	journaltitle = {Neural Computation},
	shortjournal = {Neural Computation},
	publisher = {{MIT} Press},
	author = {Dietterich, Thomas G.},
	urldate = {2026-05-28},
	date = {1998-10-01},
}

@article{benavoli_time_2017,
	title = {Time for a Change: a Tutorial for Comparing Multiple Classifiers Through Bayesian Analysis},
	volume = {18},
	issn = {1533-7928},
	url = {http://jmlr.org/papers/v18/16-305.html},
	shorttitle = {Time for a Change},
	pages = {1--36},
	number = {77},
	journaltitle = {Journal of Machine Learning Research},
	author = {Benavoli, Alessio and Corani, Giorgio and Demšar, Janez and Zaffalon, Marco},
	urldate = {2026-05-28},
	date = {2017},
}

@article{nadeau_inference_2003,
	title = {Inference for the Generalization Error},
	volume = {52},
	issn = {1573-0565},
	url = {https://doi.org/10.1023/A:1024068626366},
	doi = {10.1023/A:1024068626366},
	pages = {239--281},
	number = {3},
	journaltitle = {Machine Learning},
	shortjournal = {Machine Learning},
	author = {Nadeau, Claude and Bengio, Yoshua},
	urldate = {2026-05-28},
	date = {2003-09-01},
	langid = {english},
}

\end{document}